\pdfoutput=1
\documentclass[11pt]{article}
\usepackage[preprint]{acl}
\usepackage{times}
\usepackage{latexsym}
\usepackage[T1]{fontenc}
\usepackage[utf8]{inputenc}
\usepackage{microtype}
\usepackage{inconsolata}
\usepackage{graphicx}
\usepackage[export]{adjustbox}
\usepackage{float}
\usepackage{subfigure}
\usepackage{parskip}
\usepackage{stfloats}
\usepackage{amsmath}
\usepackage{enumitem}
\usepackage{multirow}

\renewcommand{\dblfloatpagefraction}{.9}

\title{LOCR: Location-Guided Transformer for Optical Character Recognition}

\author{
Yu Sun$^{1,2}$\thanks{This work was done during her internship at Shanghai Artificial Intelligence Laboratory.} \quad Dongzhan Zhou$^{1}$ \quad Chen Lin$^3$ \quad Conghui He$^{1}$ \quad Wanli Ouyang$^{1}$ \quad Han-Sen Zhong$^{1}$ \thanks{Corresponding author: zhonghansen@pjlab.org.cn}\\ \\
$^{1}$Shanghai Artificial Intelligence Laboratory, Shanghai, 200232, China. \\ \quad $^{2}$Shanghai Jiaotong University, Shanghai, 200030, China. \\ \quad $^{3}$University of Oxford, Oxford, OX1 2JD, United Kingdom.
}

\begin{document}
\maketitle
\begin{figure*}[h]
    \centering 
    \subfigure[\textbf{Data}: dataset \& data engine]{
        \label{Fig.sub.1}
        \includegraphics[width=0.32\textwidth]{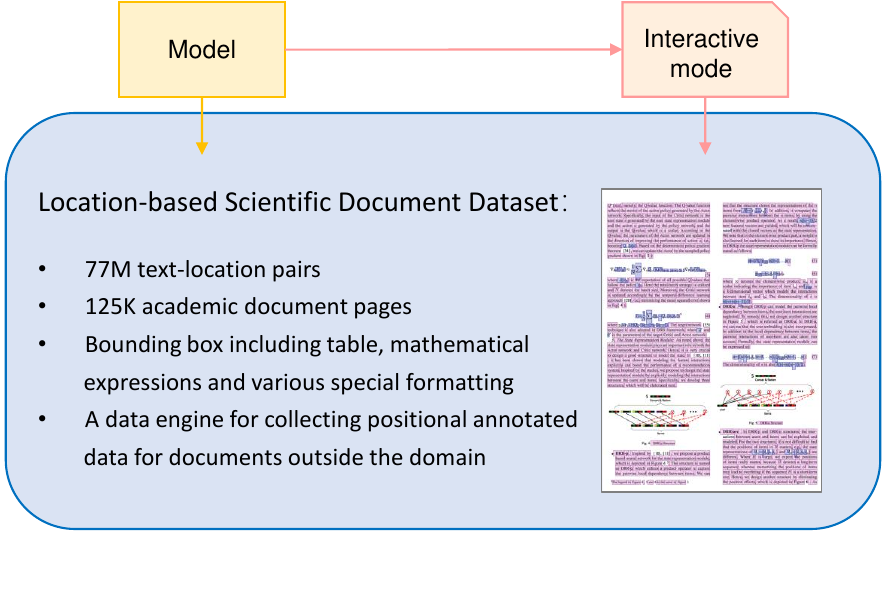}}\hspace{1mm}\subfigure[\textbf{Model}: location-guided transformer]{
        \label{Fig.sub.2}
    \includegraphics[width=0.32\textwidth]{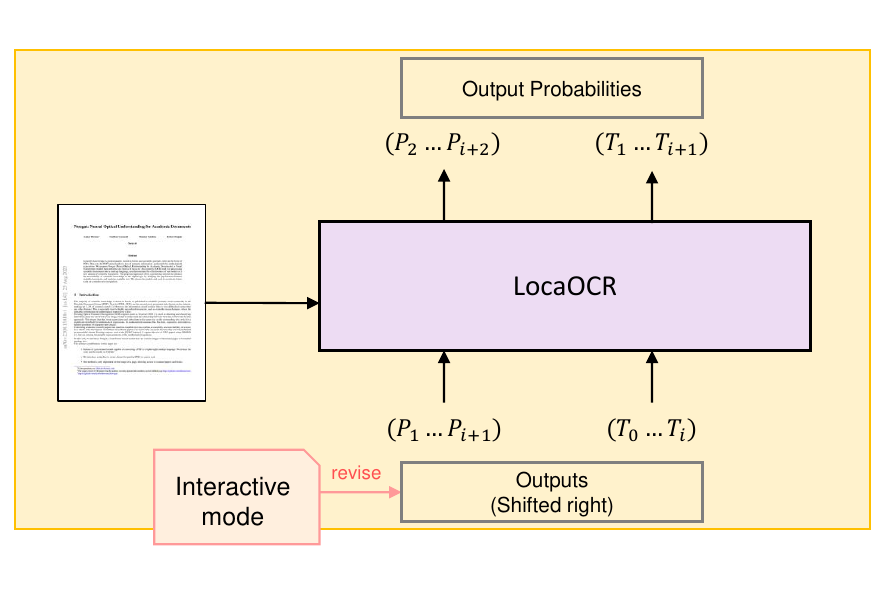}}\hspace{1mm}\subfigure[\textbf{Interactive}: align with human intent]{
        \label{Fig.sub.2}
    \includegraphics[width=0.32\textwidth]{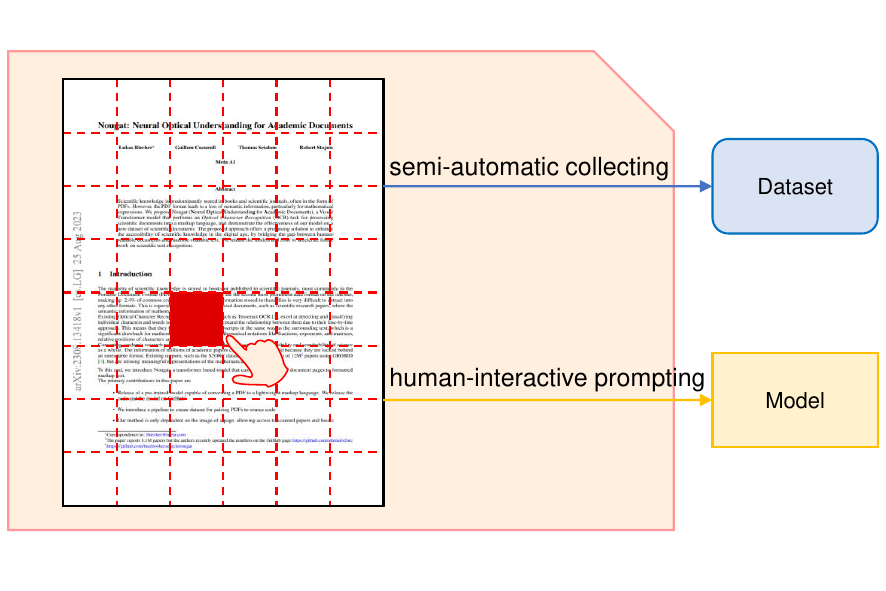}}
    \caption{An overview of three components of our work: a {\em \textbf{large-scale dataset}} with positional annotation and a data engine, a {\em \textbf{location-guided OCR model}} for various layouts, and an {\em \textbf{interactive mode}} for humans to prompt the model and modify data collection.}
    \label{fig:overview}
\end{figure*}

\begin{abstract}

Academic documents are packed with texts, equations, tables, and figures, requiring comprehensive understanding for accurate Optical Character Recognition (OCR). 
While end-to-end OCR methods offer improved accuracy over layout-based approaches, they often grapple with significant repetition issues, especially with complex layouts in Out-Of-Domain (OOD) documents.
To tackle this issue, we propose LOCR, 
a model that integrates location guiding into the transformer architecture during autoregression. We train the model on a dataset comprising over 77M text-location pairs from 125K academic document pages, including bounding boxes for words, tables and mathematical symbols. 
LOCR adeptly handles various formatting elements and generates content in Markdown language. 
It outperforms all existing methods in our test set constructed from arXiv, as measured by edit distance, BLEU, METEOR and F-measure.
LOCR also reduces repetition frequency from 4.4\% of pages to 0.5\% in the arXiv dataset, from 13.2\% to 1.3\% in OOD quantum physics documents and from 8.1\% to 1.8\% in OOD marketing documents. 
Additionally, LOCR features an interactive OCR mode, facilitating the generation of complex documents through a few location prompts from human.

\end{abstract}

\section{Introduction}

Academic literature comprises a wealth of high-quality content, yet much of it is provided in formats like PDF that are not readily for machine reading. Particularly, most academic documents of the previous centuries are scanned version. Digitizing academic documents are important for scientific research, literature retrieval, and large-language model training. However, academic document layout tends to be highly intricate, including text, equations, images, tables, and annotations, posing challenges for obtaining accurate OCR results.

One approach to document OCR is to first analyze the layout of the document and then extract the text content (\citealp{DocBed},\citealp{DocTR}).
While progress has been made in any of the two stages or handling specific types of elements, such as table detection and recognition (\citealp{yang2022tableformer}), handwritten formula recognition (\citealp{handwritten2023}) and structured information extraction (\citealp{lu2022unified}; \citealp{liao2023doctr}), it is very difficult for models to understand all the elements in an academic document and connect the different chunks into a coherent sequence.

Recently, an end-to-end transformer structure, Donut \cite{kim2022donut}, was proposed for document understanding. It effectively addresses the complexity of combining multiple models and the issue of error propagation. Without too many changes in the model, Nougat \cite{blecher2023nougat} processes academic PDFs into markup language. However, these methods are prone to hallucination and repetitions, such as incorrectly generating non-existent authors or continuously repeating the same sentence on a page.

In fact, getting trapped in a repetitive loop is a common problem with Transformer-based models sampling with greedy search decoding \citep{holtzman2019repetition}. 
It is challenging for a language model to accurately capture all the content of text-intensive documents without position perception in the autoregressive process. By visualizing the heatmap of cross-attention during the prediction process of Nougat, we found that the cross-attention cannot be focused on the correct position when the layout is complex, resulting in repetition degeneration of the model. Details of the visualized process can be found in Appendix~\ref{appendix:nougat repetition}. 
This phenomenon indicates that the positional information influence the text decoding to a great extent. We could guide the model to focus on the correct word by assigning its position.
Inspired by this, we consider incorporating positional guidance a feasible way to address the issue of repetitive loop. 
We introduce LOCR, a location-guided document understanding model, together with an original large-scale dataset and an interactive OCR mode to align with human intention (see Figure~\ref{fig:overview} for an overview).

The most significant feature that distinguishes our model from previous works is the incorporation of positional autoregression alongside text autoregression. Different from two-stage OCR, LOCR simultaneously predicts the current token and the position of the next token, which is used to prompt the decoding of the next token. Through this method, we not only combine positional information with text information but also avoid the tedious process and error accumulation in the two-stage OCR method.
Taking document images as input, our model outputs document content in Markdown format, including special formats such as superscripts and subscripts.

Furthermore, the introduction of positional supervision makes it possible to intuitively penalize locations that have already been visited by the model to avoid repetition. With the record of visited locations, we perform an importance decay strategy on positions that have been visited during the autoregressive process or are blank in the image. The repetition behavior decreases from 4.4\% of documents to 0.5\% in the arXiv test set, and from 13.2\% to 1.3\% for out-of-domain documents. For documents with complex layouts, we also introduce an interactive OCR mode. In this mode, the model would continue to decode the text where the user has dragged a box. With these enhancement strategies, the generation ability of the model is significantly improved.

Additionally, we propose a data engine for constructing academic document OCR dataset with positional annotations. We collect a large-scale dataset of 125K academic document pages with 77M text-location pairs. To the best of our knowledge, it is the first dataset that includes a bounding box of each mathematical symbol in academic documents.

In summary, the main contributions of this paper are:

\begin{itemize}[leftmargin=3mm, itemsep=0mm]
    \item We introduce LOCR, a transformer-structured OCR model with positional supervision. Our model achieves the state-of-the-art score in academic document understanding task in the arXiv test set (see Section~\ref{section:metrics}) and alleviates the repetitive degradation problem to a great extent (see Section~\ref{section:repetition}). 

    \item We innovatively introduce an interactive OCR mode, enabling the model to handle any out-of-domain documents. Humans only need to provide the position box for the next word without any cumbersome operations (see Section~\ref{section:interaction}).
    
    \item We will release a large-scale dataset composed of 125K pages of academic documents. Each piece of data contains a page image, the corresponding texts in Markdown format, and the bounding boxes of all words and mathematical symbols (see Section~\ref{section:dataset}).
\end{itemize}

\section{Related Work}

\subsection{General-purpose OCR}

Optical Character Recognition (OCR) caters to a diverse array of applications, including document digitization (\citealp{4376991}; \citealp{2017arXiv170408628M}), handwriting recognition, and scene text recognition (\citealp{2022arXiv220706966B}; \citealp{2021arXiv210407787H}; \citealp{2021arXiv210910282L}). The classic OCR methods consist of two stages: text detection and text recognition. The text detection algorithm obtains the position of text boxes from the image, and then the recognition algorithm recognizes the content within the text boxes. Researches in these sub-fields have achieved satisfactory results, such as EAST \cite{EAST} for text detection, CRNN \cite{CRNN} for text recognition, and LayoutLM family (\citealp{LayoutLM-v1}; \citealp{LayoutLM-v2}; \citealp{LayoutLM-v3}) for document element identification. There also has been various integrated toolbox to connect the above functions, such as DocXChain \cite{DocXChain} and EffOCR \cite{EfficientOCR}.

\subsection{Academic document OCR}

For academic document understanding, additional tasks like table and mathematical equation parsing are also involved. Marker \cite{marker} is a pipeline of text extracting, layout detection, and block combination, which converts PDF, EPUB, and MOBI to Markdown with a series of deep learning models. PaddleOCR develops an intelligent document analysis system  PP-Structure \cite{PaddleOCR}, which first identifies the direction of an image and then completes two tasks of layout information analysis and key information extraction. 
Such OCR-based approaches have shown promising performance but suffer from complexity and error propagation to the subsequent process. To address this issue, document understanding models based on transformer structure were proposed. Donut \cite{kim2022donut} is an encoder-decoder model that directly decodes the expected sequences from visual inputs. Nougat \cite{blecher2023nougat} is a specific model trained on academic documents to process academic PDFs into markup language. It combines an image encoder and a token decoder, with the ability to parse images of math equations and tables. 

With the emergence of general large models, some Large Vision-Language Models (LVLMs) mark a significant milestone across a range of OCR tasks. MEGVII proposed Vary \cite{wei2023vary}, a document parsing method by scaling up the vision vocabulary of LVLMs, equipping the large model with the fine-grained perception and understanding. As the state-of-the-art multimodality model, GPT-4v \cite{GPT-4v} performs well in recognizing and understanding Latin contents. But it shows limitations when dealing with complex tasks such as table structure recognition and end-to-end semantic entity recognition \citep{shi2023exploring}. When it comes to unstructured layouts or inconsistent text distribution, GPT-4v tends to omit lengthy tables and only reconstruct the short beginning of that.

Without the box detection of two-stage OCR, such methods are prone to hallucination and repetitions. This phenomenon indicates that it is crucial for the model to find the correct position in order to generate the correct sequences, especially for ambiguous layouts and out-of-domain documents.

\subsection{Promptable model}

Interactive models play a significant role in aligning behavior of artifical intelligence with human intentions, which have shown promising performance within a variety of domains. SAM\cite{kirillov2023segment} presents an interactive segmentation model capable of accommodating point, box, and text-based input. DINOv \cite{DINOv} achieves visual in-context prompting in both referring and general segmentation. T-Rex \cite{T-Rex} explores object detection and counting, which can interactively refine the counting results by prompting on missing or falsely-detected objects.
In contrast, the field of OCR revolves less interactive explorations, despite the dealing with complex layout has an urge for human prompts and interactions.

\section{Dataset}\label{section:dataset}

\subsection{Data collection}

\begin{figure*}[htbp]
    \centering
    \includegraphics[width=1\textwidth]{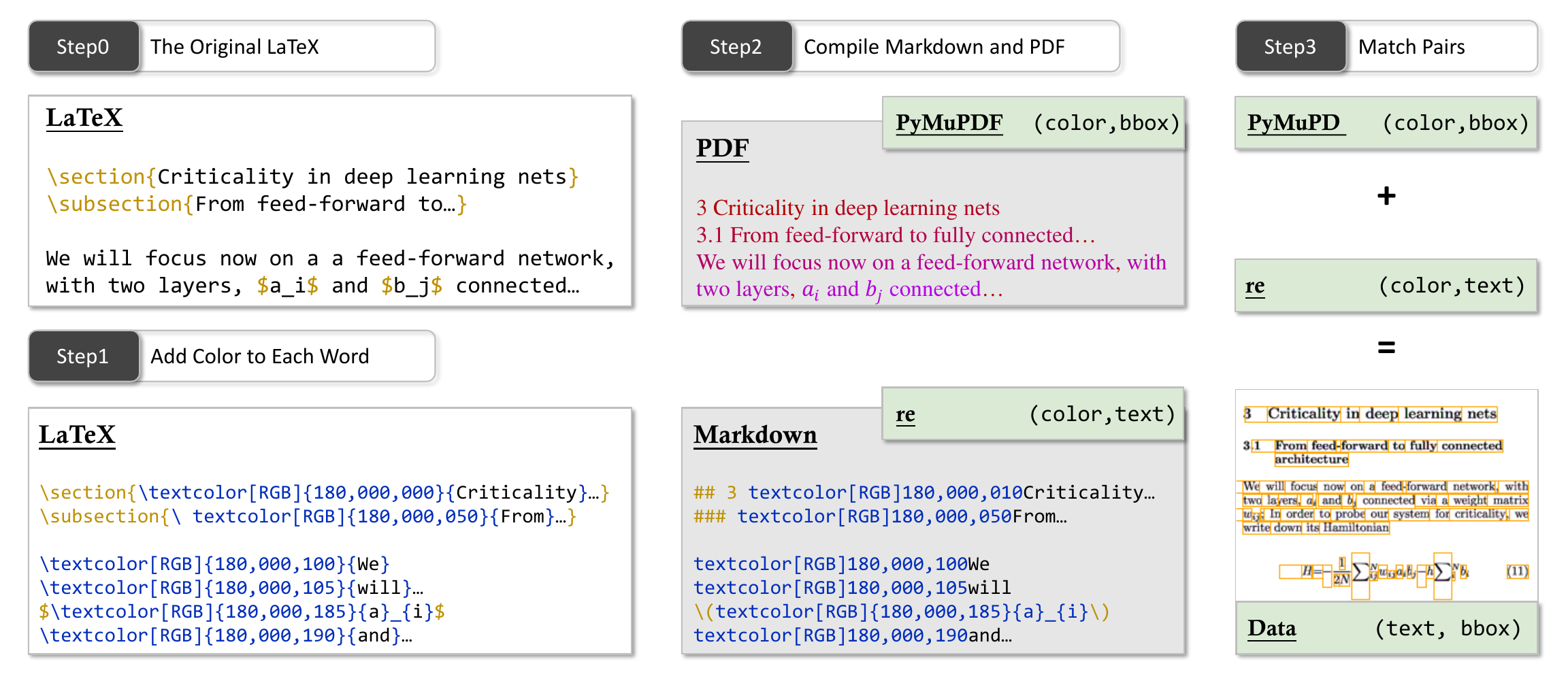}
    \caption{Data Processing. Step1: Add a unique RGB identifier to each word by parsing the Tex file. Step2: Convert source file into Markdown and PDF formats respectively. Step3: Extract color-bbox pairs from colored PDF, color-text pairs from Markdown, and merge the two to get the text-bbox pairs.}
    \label{fig:Data_Processing}
\end{figure*}

To the best of our knowledge, there is no paired dataset containing markup-formatted document contents along with corresponding bounding boxes (bbox) for each word and mathematical symbol. We proposed a data engine to collect such paired data. The process is shown in Figure~\ref{fig:Data_Processing}.

We get the Tex source files of academic papers from arXiv. In the first step, we assign a unique RGB color identifier to each word or mathematical symbol automatically by using xcolor package in LaTeX (see Step1). In the second step, we follow the same pipeline as Nougat \cite{blecher2023nougat} and compile LaTeX files into PDF and Markdown files respectively. Since PDF is a rich text format that supports color changes, we obtain colorful PDF files. Meanwhile Markdown is a plain text format and the RGB identifiers are compiled into text forms (see Step2).
In the third step, we use the PyMuPDF package of python to parse the colorful PDF files and extract the pair of (color, bbox). At the same time, we parse the Markdown file with regular expressions to get the paired (color, text) data. Finally, we merge the two pairs of data by the key of RGB color to get paired (text, bbox) data (see Step3).

We collected academic papers released on arXiv from 2007 to 2023. During data processing, some articles failed the conversion due to user-defined macros or non-standardized formats. After all conversion and data cleaning, our dataset is composed of 125738 pages, which include, but are not limited to, the bounding box of plain text, Greek letters, arithmetic symbols, superscripts, subscripts, and tabular symbols. For invisible Markdown symbols like title symbols or line breaks, we assign the position of the next visible token to them. Examples of our dataset is available in Appedix~\ref{fig:Dataset}.

\subsection{Data augmentation}

\textbf{Image augmentation} To simulate the imperfections and variability of scanned documents, we follow \cite{2003transforms} to apply data augmentation to document images, including of erosion, dilation, gaussian noise, gaussian blur, bitmap conversion, image compression, grid distortion and elastic transform. Each of the transformations is applied with a certain probability.

\textbf{Text augmentation} To address the issue of the model getting stuck in repetitive loops, we randomly skip 0 to 5 tokens and their corresponding positions in the ground truth labels. Different from the perturbation method in Nougat, which randomly replaces tokens rather than skip tokens, our method shows a more pronounced effect (see Section~\ref{section:repetition}).

\textbf{Position augmentation} Since bounding boxes are involved in the autoregressive process, there may be some imprecise output. In some cases, a user may also draw a loose box in the interactive mode. Therefore, it is reasonable to add noise to the bounding boxes during the training phase. We add Gaussian noise with a mean of 0 and a standard deviation of 0.5 times the side length to each box. 

\section{Methodology}

\subsection{Model structure}

\begin{figure*}[htbp]
    \centering
    \includegraphics[width=0.8\linewidth]{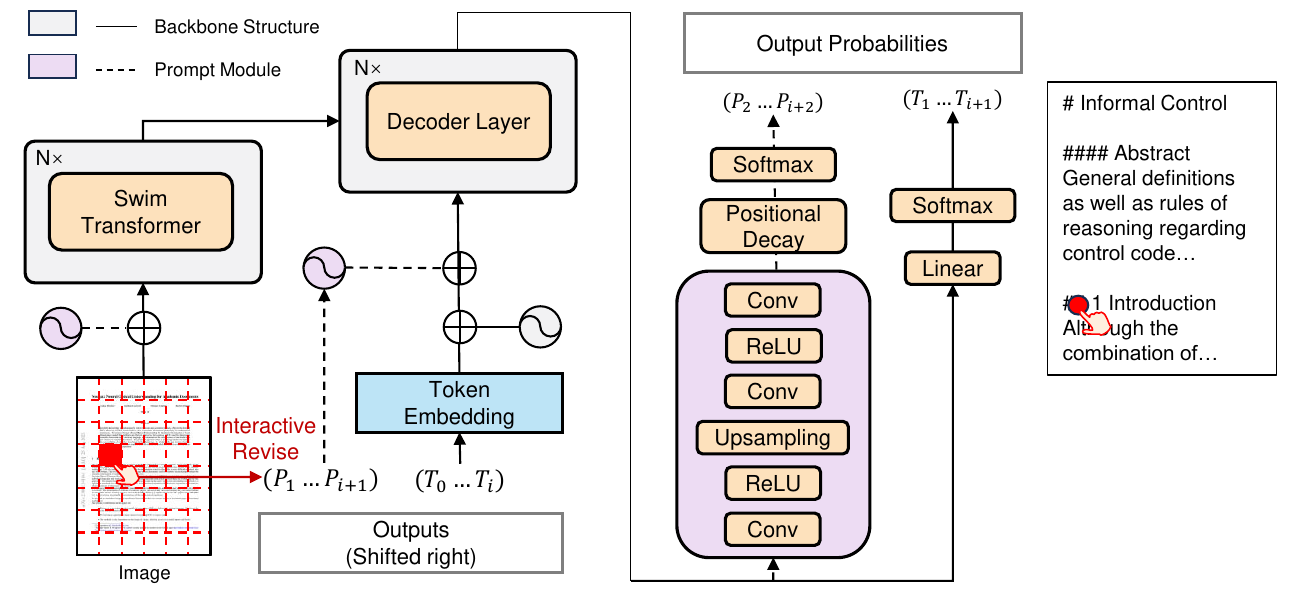}
    \caption{Model Architecture. Left: Image encoder and decoder of transformer structure. Right: Position detection head and token projection. Purple: Prompt module consisting of positional encodings and position detection head. Red: Interactive mode with human-reviewed input.}
    \label{fig:model_structure}
\end{figure*}

The over view of our model is shown in Figure~\ref{fig:model_structure}, with a transformer-based backbone and an additional prompt module to process positional information. Given an image as input, the image encoder transforms it as image embedding. Semantic information and visual information are integrated within the decoder, enabling simultaneous prediction of the current token and its next position.

\textbf{Backbone} Theoretically, our prompt module can be applied to any multimodal models with transformer structure, consisting of an image encoder and a text decoder. When no positional information is provided, the backbone model would autonomously generate sequences. In this paper, we choose Nougat \cite{blecher2023nougat} as the backbone, which uses the implementation of Swin Transformer \cite{liu2021swin} as image encoder and mBART \cite{lewis2019bart} as decoder. Given an image of $x\in R^{3,H_0,W_0}$, the image encoder transfers it into dense embedding $h_{img} \in R^{H,W,d}$, which is then decoded into a sequence of token embeddings $h_{t}\in R^{d}$. Finally, the sequence of token embeddings is projected into a logit matrix with the size of the vocabulary v.

\textbf{Prompt Module}
Without location guiding, the backbone model may get confused about where to find the next token. The prompt module is designed to perceive spatial information prompted by previous steps or human, so that the model can find the next token successfully. The prompt module consists of two-dimensional positional encoding and position detection heads. 

We opt for positional encodings with Fourier Features \cite{2020Fourier} to represent the positions of both token bounding boxes and the image. The token bounding box, defined by its top-left and bottom-right corners, is transformed into a dense position embedding $h_{box}\in R^{d}$. For the image embedding $h_{img} \in R^{H,W,d}$, we divide it into grids of size (H, W) (shown in Figure~\ref{fig:model_structure}), and apply positional encodings to each grid box to get the its position embedding $h_{grid} \in R^{H,W,d}$. 

The position detection heads are used to predict the position of the next token. Given that the weights of the cross-attention layers indicate the similarity between image grids and the current token, we utilize them as input for position detection. Inspired by CenterNet \citep{duan2019centernet}, an effective object detection algorithm, we use three convolutional heads with similar structure to predict the position of a token. The first convolution head predicts a heatmap of size (H,W) and conducts a classification task with softmax to find the grid containing the next token. The second and third convolution heads regress the size and center offset of the next bounding box respectively. Finally, the coordinates of the bounding box are calculated based on the center point and the width and height. To improve prediction accuracy, we upsample the image grid output by decoder from (H,W) to (2H,2W), allowing finer-grained positition prediction.

\textbf{Information fusion}
The token information and spatial information is fused in cross-attention layers of decoder. In backbone models without prompt module, the cross-attention layers take solely image embedding as encoder hidden states and solely token embedding as hidden states input. Instead, we use the sum of the image embedding $h_{img} \in R^{H,W,d}$ and its position embedding $h_{grid}\in R^{H,W,d}$ as the encoder hidden states, and the sum of token embedding $h_{t}\in R^{d}$ and position embedding $h_{box}\in R^{d}$ as the hidden states input. As a consequence, in cross-attention layers where token information interacts with the image contents, the positional information of tokens also interacts with that of the image. 

\subsection{Decay strategy for anti-repetition}

During the inference stage, we introduce position decay strategies based on prior knowledge to guide the prediction of positions.

\textbf{Accumulation Decay} The core of the accumulation decay strategy is to record the count of tokens that have appeared in each grid. When the position detection head predicts subsequent positions, grids where many tokens have already been located will be penalized with a decay rate. 
The heatmap for predicting the next grid is adjusted as follows:
\begin{equation}
hm = hm + log(\sigma) \cdot cnt 
\end{equation}

Where $hm\in R^{2H,2W}$ denotes the upsampled heatmap predicted by the first position detection head and $cnt \in R^{2H,2W}$ denotes the count of tokens that have appeared in each grid.
The $\sigma \in (0,1]$ denotes decay rate and $cnt$ is the accumulative counts. When $\sigma$ is set to 1, the decay function is deactivated. Smaller $\sigma$ value means stronger decay effect. We recommend using a decay rate between 0.75 and 0.95, depending on the density of text in the target documents and the formatting style.

\textbf{Blank Decay}
Another intuitive idea is to apply positional decay to blank grids. We calculate the standard deviation for pixels within each grid, where grids with smaller standard deviations (in extreme cases, containing no characters at all) are considered less likely to contain the next token. Together with blank decay strategy, the heatmap is adjusted as follows:

\begin{equation}
hm = hm + log(\sigma) \cdot cnt + log(\eta \cdot std)
\end{equation}

\subsection{Loss function}

Our loss function consists of two parts: token loss and position loss.

\textbf{Token loss} We use the cross-entropy loss of tokens to train the language decoder. 

\textbf{Position loss} For the three convolutional heads in the position detection module, we apply cross-entropy loss to the first classification head and supervise the subsequent two heads using the Intersection over Union (IOU) metric. Additionally, we integrate the normalized Euclidean distance between the center of the predicted box and that of the target box to mitigate the shortcomings of slow convergence and inaccurate regression inherent in IOU \cite{2019diou}. The position loss function is as follows:

\begin{equation}
    L_{p} = \alpha L_{p}^{ce}+\beta (1-iou+\gamma {d}^2)
\end{equation}

As the prediction of the text at the beginning of a page is much more challenging and important, we assigned a higher weight $\theta$ for the initial text than the subsequent text.

The final loss function is as follows:

\begin{equation}
    l = \theta (L_p^{init}+L_{t}^{init}) + L_p^{sub}  + L_t^{sub}
\end{equation}

\subsection{Human interaction}

As a complement to our location-guided OCR method, we provide an interactive mode, which serves both for improving the model's performance and as a part of our data construction engine.

\textbf{Model Assistant} In the interactive mode, We provide a browser-based tool to enable users to give real-time position prompts by simply dragging a box. LOCR takes the $i$-th token and the $(i+1)$-th position as inputs, simultaneously predicting the $(i+1)$-th token and the $(i+2)$-th position. When the autoregressive process encounters a state of confusion, characterized by a predicted token or position confidence lower than a predetermined threshold, users can opt to provide a positional prompt. With the correct position provided, the autoregressive process would go on more smoothly (see Section~\ref{section:interaction} for results).

\textbf{Data construction} With the model automatically predicting positions, minimal human intervention is required to acquire additional out-of-domain data. Our positional encoding and detection modules can smoothly convert the bounding box between human-readable coordinate representations and machine-friendly dense embedding, making the idea easy to implement. This paves the way for broader applications of location-based OCR method.

\section{Result and Evaluation}\label{Result and Evaluation}

\begin{figure*}
    \centering 
    \subfigure{
        \label{Fig.sub.1}
        \includegraphics[width=0.5\textwidth,height=0.7\textwidth,frame,trim={1.5cm 2cm 1.5cm 2cm},clip]{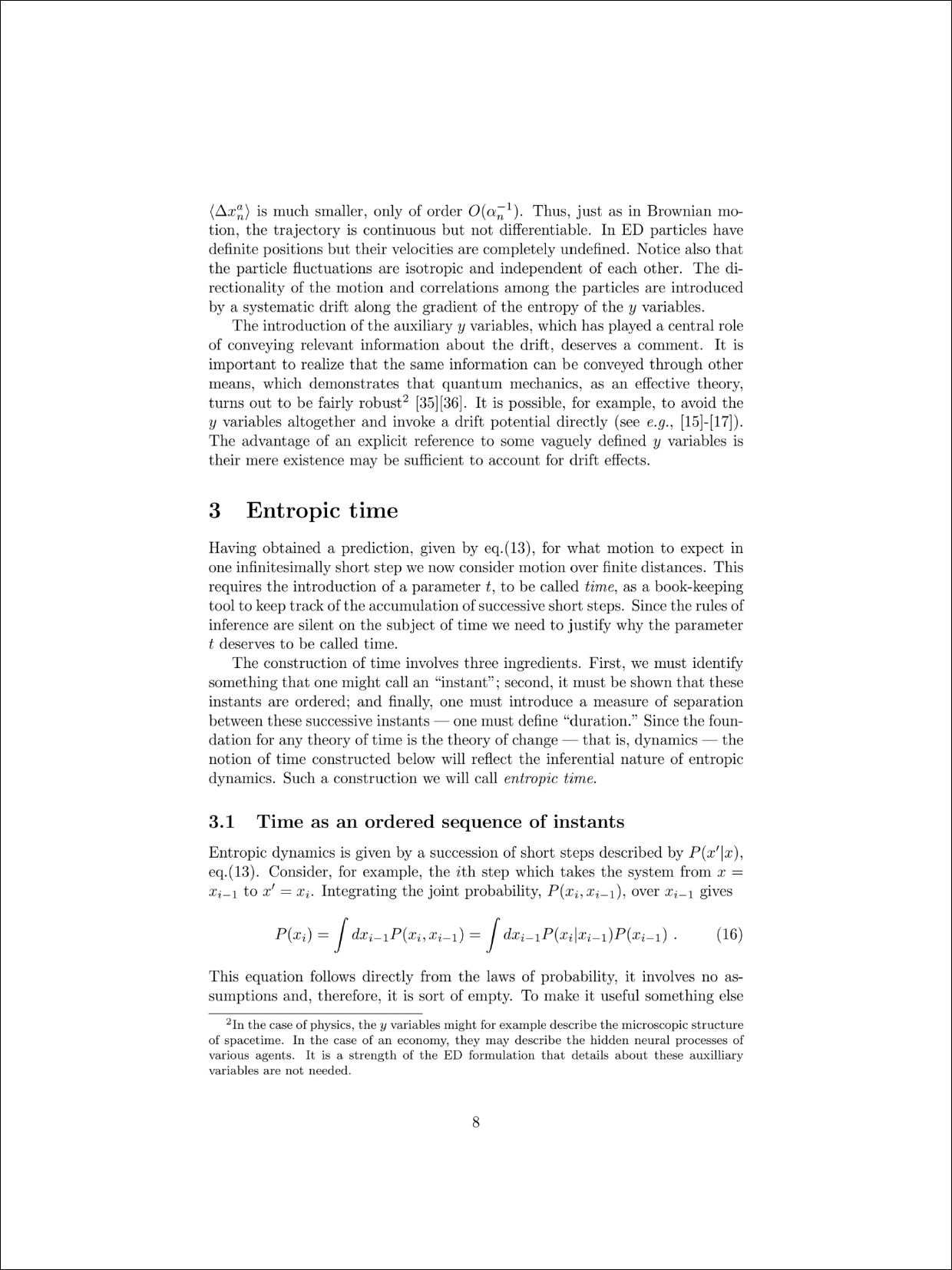}}\subfigure{
        \label{Fig.sub.2}
        \includegraphics[width=0.5\textwidth,height=0.7\textwidth,frame,trim={1.5cm 2cm 1.5cm 2cm},clip]{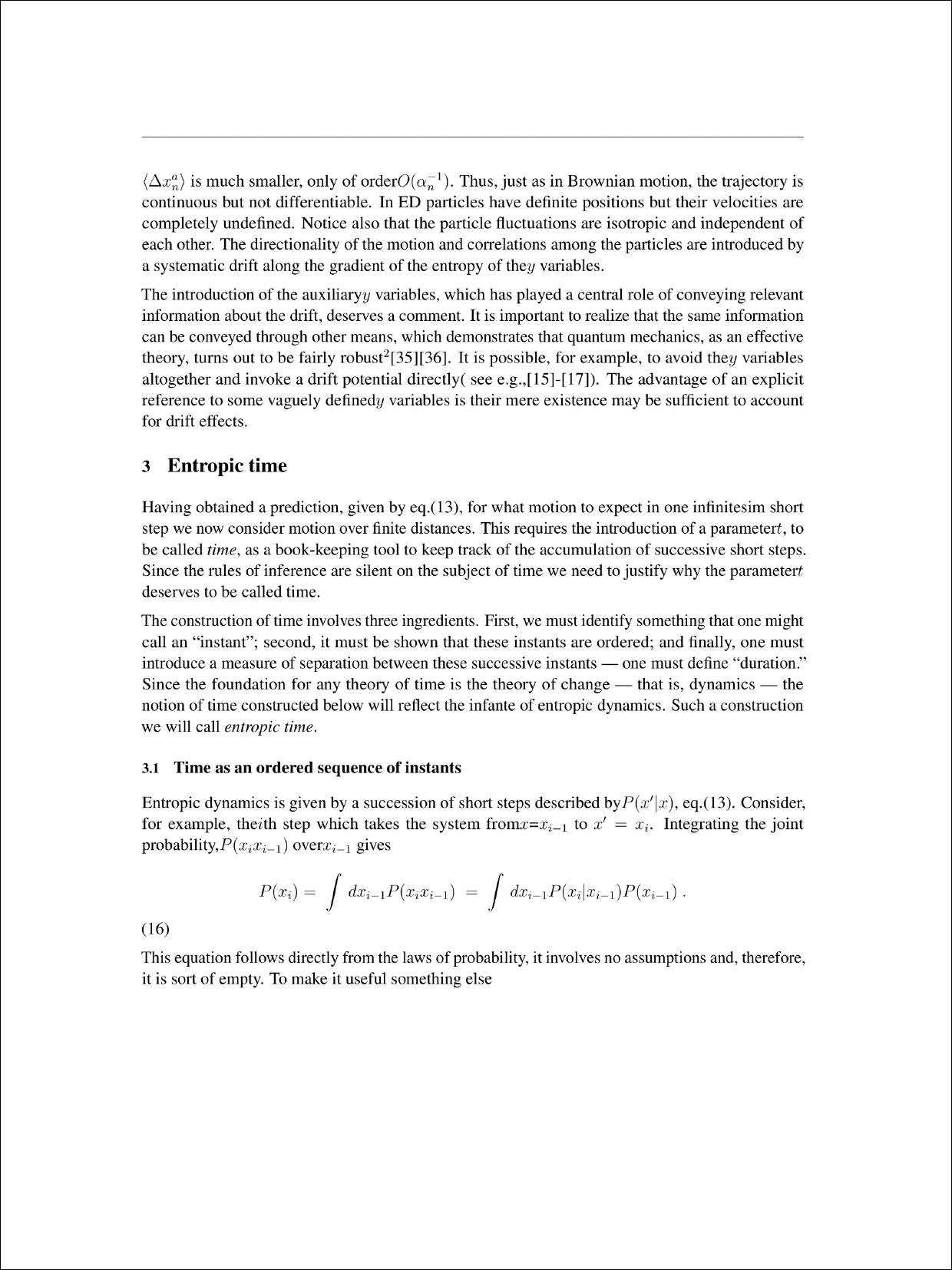}}
    \caption{Examples of our model output. Left: Origin image of document page. Right: Model output converted to Markdown and rendered back into a PDF. More detailed examples are available in Appendix~\ref{appendix:examples}}
    \label{fig:mmd examples}
\end{figure*}

\subsection{Implementation details}

\textbf{Baseline}
We use both the state-of-the-art integrated toolbox Marker, PaddleOCR and end-to-end generation model Nougat as our baselines. For PaddleOCR, which outputs each bounding box by text detection and corresponding text by text recognition, we concatenate the sequences in the order of its model output.

\textbf{Dataset} Since our main baseline model, Nougat, does not provide an open resource dataset, we evaluate our method with the dataset introduced in Section~\ref{section:dataset}, which shares the same data source and processing pipeline as Nougat. The test set contains 1000 pages of academic documents and each piece of data consists of a triplet (image, text, bounding box). In the testing phase, only images are used as inputs, while the text and bounding boxes serve solely for evaluating model performance. Therefore, our evaluation method is fair and reasonable.

\textbf{Setup} We resize the input dimensions of the images to ($H_0$, $W_0$) = (896, 672), an aspect ratio that accommodates the majority of academic paper sizes. 
The maximal sequence length of transformer decoder is set to 4096 to allow the output of intensive text in academic research papers. During inference the text is generated using greedy decoding.

\textbf{Training details} We initialize the backbone parameters using the pretrained Nougat small model, while the prompt module is initialized randomly. Our model has 248~M parameters and was trained for three days using 128 A100 80GB GPUs, with a total batch size of 256. The maximum learning rate is set to $5\times 10^{-4}$, with exponential decay until reaching $1\times 10^{-5}$.

\subsection{Metrics}\label{section:metrics}

\begin{table*}[hbt]
    \centering
    \begin{tabular}{c|cccccc}
        \hline
        \textbf{Method} & \textbf{Edit dist$\downarrow$} & \textbf{BLEU$\uparrow$} & \textbf{METEOR$\uparrow$} & \textbf{Precision$\uparrow$} & \textbf{Recall$\uparrow$} & \textbf{F1$\uparrow$}\\
        \hline
        PaddleOCR & 0.475 & 0.500 & 0.589 & 0.713 & 0.690 & 0.696 \\
        Marker & 0.221 & 0.696 & 0.783 & 0.838 & 0.804 & 0.814 \\
        \hline
        Nougat small (247M*) & 0.209 & 0.789 & 0.851 & 0.887 & 0.874 & 0.867 \\
        Nougat base (348M*) & 0.201 & 0.801 & 0.856 & 0.893 & 0.880 & 0.876 \\
        \hline
        LOCR (248M*, $\sigma=1$) & 0.153 & 0.786 & 0.864 & 0.890 & 0.871 & 0.880\\
        LOCR ($\sigma=0.85$) & \textbf{0.125} & \textbf{0.827} & \textbf{0.893} & \textbf{0.898} & \textbf{0.897} & \textbf{0.897}\\
        LOCR ($\sigma=0.75$) & 0.127 & 0.824 & 0.890 & 0.895 & 0.894 & 0.894\\
        \hline
    \end{tabular}
    \caption{\label{metrics} Comparative performance results on the arXiv test set. Our LOCR method demonstrates superior performance across multiple metrics, significantly outperforming the baseline methods. Notably, LOCR with $\sigma = 0.85$ shows the best overall balance of high precision, recall, and F1 scores, along with the lowest edit distance and the highest BLEU and METEOR scores, confirming the effectiveness of our approach, especially when positional decay is finely tuned ($\sigma = 0.85$). The first entry for LOCR indicates performance without positional decay, illustrating the impact of this feature on the model's accuracy. *Number of parameters.    }
\end{table*}

Following Nougat \cite{blecher2023nougat}, we use Edit distance, BLEU \cite{bleu}, METEOR \cite{meteor}, Precision, Recall and F-measure to characterize the quality of output text. 

As shown in Table~\ref{metrics}, while the number of LOCR's parameters is only slightly more than the small version of Nougat, our model outperforms the base version of Nougat in all evaluation metrics when the decay rate is set to 0.85. In contrast, Marker, as a multi-stage pipeline, dose not convert all equations to LaTeX and not all lines are joined properly. For the autogressive method without position supervision, Nougat prones to hallucination and repetition. These results firmly demonstrate the effectiveness of LOCR model and the positional decay strategy.

Besides, we use IOU metrics to measure the performance of our prompt module. LOCR achieves a IOU score of 0.702. Our method successfully handles various layouts, including pages with multiple subfigures, tables, mathematical formulas, and references (Examples are available in Appendix~\ref{appendix:examples}).

\subsection{Repetition}\label{section:repetition}

\begin{table*}[hbt]
    \centering
    {
    \begin{tabular}{c|ccc|ccc|ccc}
        \hline
         \multirow{2}{*}{\textbf{Method}} & \multicolumn{3}{c|}{\textbf{ArXiv}} & \multicolumn{3}{c|}{\textbf{Quantum}} &  \multicolumn{3}{c}{\textbf{Marketing}} \\
        & Page & Doc* & Cover & Page & Doc* & Cover & Page & Doc* & Cover  \\
        \hline
        Nougat small & 4.4\% & 27.6\% & 6.4\% & 13.8\% & 63.9\% & 22.6\% & 8.3\% & 60.8\% & 14.5\% \\
        Nougat base & 4.4\%  & 27.8\% & 5.3\% & 13.2\% & 55.4\% & 15.4\% & 8.1\% & 60.2\% & 16.9\% \\
        \hline
        LOCR ($\sigma=1$) & 7.5\%  & 37.3\% & 3.3\% & 18.1\% & 69.6\% & 6.9\% & 8.8\% & 50.3\% & 8.3\% \\
        LOCR ($\sigma=0.85$) & 0.6\% & 3.6\% & 0.3\% & \textbf{1.3\%} & \textbf{9.2\%} & { 1.3\%}  &\textbf{ 1.8\%} & \textbf{17.4\%} &\textbf{1.5\%}\\
        LOCR ($\sigma=0.75$) & \textbf{0.5\%} & \textbf{3.2\%} & \textbf{0.1\%} & 1.5\% & 10.9\% & \textbf{0.7\%}  & 1.9\% & 17.5\% & 1.9\% \\
        \hline
    \end{tabular}
    }
    \caption{
    \label{repetition metrics} Robustness of LOCR across diverse domains, showcasing the significant reduction in generation failures with our LOCR model and decay strategy, especially for complex layouts such as the cover. 
    The three columns for each domain are calculated based on failed pages / total pages, failed doc / total doc, and doc with failed cover / total doc.
    *Statistics on the number of pages corresponding to each document in the test set can be found in Appendix~\ref{appendix:statistics of test documents}.
    }
\end{table*}

Following Nougat \cite{blecher2023nougat}, we detect the repetition behavior during inference by computing the variances of the largest logit values of each step. If the signal drops below a certain threshold from a certain token, we regard the sequence to have repetitions.

We evaluate the generation ability of our model and present the frequency of repetitive degeneration in Table~\ref{repetition metrics}. To cover as much subject content and layout as possible, we selected 1000 papers each from natural sciences (quantum physics)  and social sciences (marketing), as out-of-domain test documents.
We calculate both the proportion of failed pages and that of failed documents. As the first page of an academic document typically shows a more complex layout than the subsequent pages, we additionally calculate the proportion of documents with failures in the cover. The model exhibits an impressive decrease in repetition-induced failures, achieving a substantial improvement over the Nougat base across the arXiv, quantum, and marketing test sets, indicating a marked increase in reliability and accuracy in document generation tasks. 

Specifically, in arXiv dataset, LOCR with $\sigma = 0.75$ reduces the failure rate for all pages from 4.4\% to 0.5\%. For OOD documents, where the document content is more challenging to comprehend with longer and more complex formulas, LOCR with $\sigma = 0.85$ reduces the failure rate for all pages to 1.3\% for quantum documents and 1.8\% for marketing documents.
On the other hand, among all failed documents, the proportion of failures on the first page is significantly decreased, demonstrating better ability of LOCR to handle more complex layouts. These results underscore the efficacy of our model in handling complex document structures with a high degree of success. Some pages that failed with Nougat but were successfully converted by LOCR are shown in Appendix~\ref{appendix:examples}.

\subsection{Interaction}\label{section:interaction}

Although the problem of repetitive degeneration has been largely alleviated, we aim to complete the remaining layouts in the interactive mode. When the model encounters a layout that is difficult to judge and the confidence of the predicted position is lower than the threshold, simply dragging a bounding box allows the model to automatically return to the expected position and continue outputting correct results. Interactive examples are available in {Appendix~\ref{appendix:interaction mode}}

\section{Discussion}

In document OCR, each generated token corresponds to a specific location in the input image. In our work, we introduce LOCR, which incorporates location guiding into the language model to enhance the performance of OCR tasks. Moreover, our approach significantly mitigates the problem of repetitive loops often encountered by transformer-based models during greedy search. LOCR also allows for interactive correction in cases of errors or low confidence outputs, particularly when dealing with OOD complex layouts. Users can prompt the location interactively, guiding the model generates accurate OCR results.

We believe that LOCR and interactive tool can be applied to digitize documents from various fields with complex layouts, thereby assisting academic research, literature retrieval, and large language model training. Furthermore, the OCR datasets with location guiding can facilitate the community develops better OCR models. In turn, the interactive semi-automatic data engine can be utilized to construct datasets for fine-tuning OCR models to specific domain literature, and enhancing the generalization capability of our model. We hope this work can help the development of the area of OCR.



\section*{Acknowledgements}
We sincerely thank Xingyu Zeng and Guoqiang Jin for their invaluable discussion and constructive feedback. This work is partially supported by the National Key R\&D Program of China (NO.2022ZD0160100), and in part by Shanghai Committee of Science and Technology (Grant No. 21DZ1100100).

\bibliography{custom}

\begin{thebibliography}{37}
\expandafter\ifx\csname natexlab\endcsname\relax\def\natexlab#1{#1}\fi

\bibitem[{Banerjee and Lavie(2005)}]{meteor}
Satanjeev Banerjee and Alon Lavie. 2005.
\newblock Meteor: An automatic metric for mt evaluation with improved correlation with human judgments.
\newblock In \emph{Proceedings of the acl workshop on intrinsic and extrinsic evaluation measures for machine translation and/or summarization}, pages 65--72.

\bibitem[{{Bautista} and {Atienza}(2022)}]{2022arXiv220706966B}
Darwin {Bautista} and Rowel {Atienza}. 2022.
\newblock \href {https://doi.org/10.48550/arXiv.2207.06966} {{Scene Text Recognition with Permuted Autoregressive Sequence Models}}.
\newblock \emph{arXiv e-prints}, page arXiv:2207.06966.

\bibitem[{Blecher et~al.(2023)Blecher, Cucurull, Scialom, and Stojnic}]{blecher2023nougat}
Lukas Blecher, Guillem Cucurull, Thomas Scialom, and Robert Stojnic. 2023.
\newblock Nougat: Neural optical understanding for academic documents.
\newblock \emph{arXiv preprint arXiv:2308.13418}.

\bibitem[{{Bryan} et~al.(2023){Bryan}, {Carlson}, {Arora}, and {Dell}}]{EfficientOCR}
Tom {Bryan}, Jacob {Carlson}, Abhishek {Arora}, and Melissa {Dell}. 2023.
\newblock \href {https://doi.org/10.48550/arXiv.2310.10050} {{EfficientOCR: An Extensible, Open-Source Package for Efficiently Digitizing World Knowledge}}.
\newblock \emph{arXiv e-prints}, page arXiv:2310.10050.

\bibitem[{Duan et~al.(2019)Duan, Bai, Xie, Qi, Huang, and Tian}]{duan2019centernet}
Kaiwen Duan, Song Bai, Lingxi Xie, Honggang Qi, Qingming Huang, and Qi~Tian. 2019.
\newblock Centernet: Keypoint triplets for object detection.
\newblock In \emph{Proceedings of the IEEE/CVF international conference on computer vision}, pages 6569--6578.

\bibitem[{{Hernandez Diaz} et~al.(2021){Hernandez Diaz}, {Qin}, {Ingle}, {Fujii}, and {Bissacco}}]{2021arXiv210407787H}
Daniel {Hernandez Diaz}, Siyang {Qin}, Reeve {Ingle}, Yasuhisa {Fujii}, and Alessandro {Bissacco}. 2021.
\newblock \href {https://doi.org/10.48550/arXiv.2104.07787} {{Rethinking Text Line Recognition Models}}.
\newblock \emph{arXiv e-prints}, page arXiv:2104.07787.

\bibitem[{Holtzman et~al.(2019)Holtzman, Buys, Du, Forbes, and Choi}]{holtzman2019repetition}
Ari Holtzman, Jan Buys, Li~Du, Maxwell Forbes, and Yejin Choi. 2019.
\newblock The curious case of neural text degeneration.
\newblock \emph{arXiv preprint arXiv:1904.09751}.

\bibitem[{{Huang} et~al.(2022){Huang}, {Lv}, {Cui}, {Lu}, and {Wei}}]{LayoutLM-v3}
Yupan {Huang}, Tengchao {Lv}, Lei {Cui}, Yutong {Lu}, and Furu {Wei}. 2022.
\newblock \href {https://doi.org/10.48550/arXiv.2204.08387} {{LayoutLMv3: Pre-training for Document AI with Unified Text and Image Masking}}.
\newblock \emph{arXiv e-prints}, page arXiv:2204.08387.

\bibitem[{Jiang et~al.(2023)Jiang, Li, Ren, Liu, Zeng, Yu, and Zhang}]{T-Rex}
Qing Jiang, Feng Li, Tianhe Ren, Shilong Liu, Zhaoyang Zeng, Kent Yu, and Lei Zhang. 2023.
\newblock T-rex: Counting by visual prompting.
\newblock \emph{arXiv preprint arXiv:2311.13596}.

\bibitem[{Kim et~al.(2022)Kim, Hong, Yim, Nam, Park, Yim, Hwang, Yun, Han, and Park}]{kim2022donut}
Geewook Kim, Teakgyu Hong, Moonbin Yim, JeongYeon Nam, Jinyoung Park, Jinyeong Yim, Wonseok Hwang, Sangdoo Yun, Dongyoon Han, and Seunghyun Park. 2022.
\newblock Ocr-free document understanding transformer.
\newblock In \emph{European Conference on Computer Vision (ECCV)}.

\bibitem[{Kirillov et~al.(2023)Kirillov, Mintun, Ravi, Mao, Rolland, Gustafson, Xiao, Whitehead, Berg, Lo et~al.}]{kirillov2023segment}
Alexander Kirillov, Eric Mintun, Nikhila Ravi, Hanzi Mao, Chloe Rolland, Laura Gustafson, Tete Xiao, Spencer Whitehead, Alexander~C Berg, Wan-Yen Lo, et~al. 2023.
\newblock Segment anything.
\newblock \emph{arXiv preprint arXiv:2304.02643}.

\bibitem[{Lewis et~al.(2019)Lewis, Liu, Goyal, Ghazvininejad, Mohamed, Levy, Stoyanov, and Zettlemoyer}]{lewis2019bart}
Mike Lewis, Yinhan Liu, Naman Goyal, Marjan Ghazvininejad, Abdelrahman Mohamed, Omer Levy, Ves Stoyanov, and Luke Zettlemoyer. 2019.
\newblock Bart: Denoising sequence-to-sequence pre-training for natural language generation, translation, and comprehension.
\newblock \emph{arXiv preprint arXiv:1910.13461}.

\bibitem[{{Li} et~al.(2022){Li}, {Guo}, {Zhou}, {An}, {Du}, {Zhu}, {Liu}, {Hu}, and {Yu}}]{PaddleOCR}
Chenxia {Li}, Ruoyu {Guo}, Jun {Zhou}, Mengtao {An}, Yuning {Du}, Lingfeng {Zhu}, Yi~{Liu}, Xiaoguang {Hu}, and Dianhai {Yu}. 2022.
\newblock \href {https://doi.org/10.48550/arXiv.2210.05391} {{PP-StructureV2: A Stronger Document Analysis System}}.
\newblock \emph{arXiv e-prints}, page arXiv:2210.05391.

\bibitem[{Li et~al.(2023)Li, Jiang, Zhang, Ren, Liu, Zou, Xu, Li, Li, Yang et~al.}]{DINOv}
Feng Li, Qing Jiang, Hao Zhang, Tianhe Ren, Shilong Liu, Xueyan Zou, Huaizhe Xu, Hongyang Li, Chunyuan Li, Jianwei Yang, et~al. 2023.
\newblock Visual in-context prompting.
\newblock \emph{arXiv preprint arXiv:2311.13601}.

\bibitem[{{Li} et~al.(2021){Li}, {Lv}, {Chen}, {Cui}, {Lu}, {Florencio}, {Zhang}, {Li}, and {Wei}}]{2021arXiv210910282L}
Minghao {Li}, Tengchao {Lv}, Jingye {Chen}, Lei {Cui}, Yijuan {Lu}, Dinei {Florencio}, Cha {Zhang}, Zhoujun {Li}, and Furu {Wei}. 2021.
\newblock \href {https://doi.org/10.48550/arXiv.2109.10282} {{TrOCR: Transformer-based Optical Character Recognition with Pre-trained Models}}.
\newblock \emph{arXiv e-prints}, page arXiv:2109.10282.

\bibitem[{Liao et~al.(2023)Liao, RoyChowdhury, Li, Bansal, Zhang, Tu, Satzoda, Manmatha, and Mahadevan}]{liao2023doctr}
Haofu Liao, Aruni RoyChowdhury, Weijian Li, Ankan Bansal, Yuting Zhang, Zhuowen Tu, Ravi~Kumar Satzoda, R~Manmatha, and Vijay Mahadevan. 2023.
\newblock Doctr: Document transformer for structured information extraction in documents.
\newblock In \emph{Proceedings of the IEEE/CVF International Conference on Computer Vision}, pages 19584--19594.

\bibitem[{Liu et~al.(2021)Liu, Lin, Cao, Hu, Wei, Zhang, Lin, and Guo}]{liu2021swin}
Ze~Liu, Yutong Lin, Yue Cao, Han Hu, Yixuan Wei, Zheng Zhang, Stephen Lin, and Baining Guo. 2021.
\newblock Swin transformer: Hierarchical vision transformer using shifted windows.
\newblock In \emph{Proceedings of the IEEE/CVF international conference on computer vision}, pages 10012--10022.

\bibitem[{Lu et~al.(2022)Lu, Liu, Dai, Xiao, Lin, Han, Sun, and Wu}]{lu2022unified}
Yaojie Lu, Qing Liu, Dai Dai, Xinyan Xiao, Hongyu Lin, Xianpei Han, Le~Sun, and Hua Wu. 2022.
\newblock Unified structure generation for universal information extraction.
\newblock \emph{arXiv preprint arXiv:2203.12277}.

\bibitem[{mindee(2023)}]{DocTR}
mindee. 2023.
\newblock doctr: Document text recognition.
\newblock \url{https://github.com/mindee/doctr}.

\bibitem[{{Moysset} et~al.(2017){Moysset}, {Kermorvant}, and {Wolf}}]{2017arXiv170408628M}
Bastien {Moysset}, Christopher {Kermorvant}, and Christian {Wolf}. 2017.
\newblock \href {https://doi.org/10.48550/arXiv.1704.08628} {{Full-Page Text Recognition: Learning Where to Start and When to Stop}}.
\newblock \emph{arXiv e-prints}, page arXiv:1704.08628.

\bibitem[{Papineni et~al.(2002)Papineni, Roukos, Ward, and Zhu}]{bleu}
Kishore Papineni, Salim Roukos, Todd Ward, and Wei-Jing Zhu. 2002.
\newblock Bleu: a method for automatic evaluation of machine translation.
\newblock In \emph{Proceedings of the 40th annual meeting of the Association for Computational Linguistics}, pages 311--318.

\bibitem[{Paruchuri and Lampa(2023)}]{marker}
Vik Paruchuri and Samuel Lampa. 2023.
\newblock Marker: Convert pdf to markdown quickly with high accuracy.
\newblock \url{https://github.com/VikParuchuri/marker?tab=readme-ov-file}.

\bibitem[{Sakshi and Kukreja(2023)}]{handwritten2023}
Sakshi Sakshi and Vinay Kukreja. 2023.
\newblock \href {https://doi.org/10.1016/j.eswa.2022.119028} {Recent trends in mathematical expressions recognition: An lda-based analysis}.
\newblock \emph{Expert Systems with Applications}, 213:119028.

\bibitem[{{Shi} et~al.(2015){Shi}, {Bai}, and {Yao}}]{CRNN}
Baoguang {Shi}, Xiang {Bai}, and Cong {Yao}. 2015.
\newblock \href {https://doi.org/10.48550/arXiv.1507.05717} {{An End-to-End Trainable Neural Network for Image-based Sequence Recognition and Its Application to Scene Text Recognition}}.
\newblock \emph{arXiv e-prints}, page arXiv:1507.05717.

\bibitem[{Shi et~al.(2023)Shi, Peng, Liao, Lin, Chen, Liu, Zhang, and Jin}]{shi2023exploring}
Yongxin Shi, Dezhi Peng, Wenhui Liao, Zening Lin, Xinhong Chen, Chongyu Liu, Yuyi Zhang, and Lianwen Jin. 2023.
\newblock Exploring ocr capabilities of gpt-4v (ision): A quantitative and in-depth evaluation.
\newblock \emph{arXiv preprint arXiv:2310.16809}.

\bibitem[{Simard et~al.(2003)Simard, Steinkraus, and Platt}]{2003transforms}
P.Y. Simard, D.~Steinkraus, and J.C. Platt. 2003.
\newblock \href {https://doi.org/10.1109/ICDAR.2003.1227801} {Best practices for convolutional neural networks applied to visual document analysis}.
\newblock In \emph{Seventh International Conference on Document Analysis and Recognition, 2003. Proceedings.}, pages 958--963.

\bibitem[{Smith(2007)}]{4376991}
R.~Smith. 2007.
\newblock \href {https://doi.org/10.1109/ICDAR.2007.4376991} {An overview of the tesseract ocr engine}.
\newblock In \emph{Ninth International Conference on Document Analysis and Recognition (ICDAR 2007)}, volume~2, pages 629--633.

\bibitem[{{Tancik} et~al.(2020){Tancik}, {Srinivasan}, {Mildenhall}, {Fridovich-Keil}, {Raghavan}, {Singhal}, {Ramamoorthi}, {Barron}, and {Ng}}]{2020Fourier}
Matthew {Tancik}, Pratul~P. {Srinivasan}, Ben {Mildenhall}, Sara {Fridovich-Keil}, Nithin {Raghavan}, Utkarsh {Singhal}, Ravi {Ramamoorthi}, Jonathan~T. {Barron}, and Ren {Ng}. 2020.
\newblock \href {https://doi.org/10.48550/arXiv.2006.10739} {{Fourier Features Let Networks Learn High Frequency Functions in Low Dimensional Domains}}.
\newblock \emph{arXiv e-prints}, page arXiv:2006.10739.

\bibitem[{Wei et~al.(2023)Wei, Kong, Chen, Zhao, Ge, Yang, Sun, Han, and Zhang}]{wei2023vary}
Haoran Wei, Lingyu Kong, Jinyue Chen, Liang Zhao, Zheng Ge, Jinrong Yang, Jianjian Sun, Chunrui Han, and Xiangyu Zhang. 2023.
\newblock Vary: Scaling up the vision vocabulary for large vision-language models.
\newblock \emph{arXiv preprint arXiv:2312.06109}.

\bibitem[{{Xu} et~al.(2020){Xu}, {Xu}, {Lv}, {Cui}, {Wei}, {Wang}, {Lu}, {Florencio}, {Zhang}, {Che}, {Zhang}, and {Zhou}}]{LayoutLM-v2}
Yang {Xu}, Yiheng {Xu}, Tengchao {Lv}, Lei {Cui}, Furu {Wei}, Guoxin {Wang}, Yijuan {Lu}, Dinei {Florencio}, Cha {Zhang}, Wanxiang {Che}, Min {Zhang}, and Lidong {Zhou}. 2020.
\newblock \href {https://doi.org/10.48550/arXiv.2012.14740} {{LayoutLMv2: Multi-modal Pre-training for Visually-Rich Document Understanding}}.
\newblock \emph{arXiv e-prints}, page arXiv:2012.14740.

\bibitem[{{Xu} et~al.(2019){Xu}, {Li}, {Cui}, {Huang}, {Wei}, and {Zhou}}]{LayoutLM-v1}
Yiheng {Xu}, Minghao {Li}, Lei {Cui}, Shaohan {Huang}, Furu {Wei}, and Ming {Zhou}. 2019.
\newblock \href {https://doi.org/10.48550/arXiv.1912.13318} {{LayoutLM: Pre-training of Text and Layout for Document Image Understanding}}.
\newblock \emph{arXiv e-prints}, page arXiv:1912.13318.

\bibitem[{Yang et~al.(2022)Yang, Gupta, Upadhyay, He, Goel, and Paul}]{yang2022tableformer}
Jingfeng Yang, Aditya Gupta, Shyam Upadhyay, Luheng He, Rahul Goel, and Shachi Paul. 2022.
\newblock Tableformer: Robust transformer modeling for table-text encoding.
\newblock \emph{arXiv preprint arXiv:2203.00274}.

\bibitem[{{Yang} et~al.(2023){Yang}, {Li}, {Lin}, {Wang}, {Lin}, {Liu}, and {Wang}}]{GPT-4v}
Zhengyuan {Yang}, Linjie {Li}, Kevin {Lin}, Jianfeng {Wang}, Chung-Ching {Lin}, Zicheng {Liu}, and Lijuan {Wang}. 2023.
\newblock \href {https://doi.org/10.48550/arXiv.2309.17421} {{The Dawn of LMMs: Preliminary Explorations with GPT-4V(ision)}}.
\newblock \emph{arXiv e-prints}, page arXiv:2309.17421.

\bibitem[{{Yao}(2023)}]{DocXChain}
Cong {Yao}. 2023.
\newblock \href {https://doi.org/10.48550/arXiv.2310.12430} {{DocXChain: A Powerful Open-Source Toolchain for Document Parsing and Beyond}}.
\newblock \emph{arXiv e-prints}, page arXiv:2310.12430.

\bibitem[{{Zheng} et~al.(2019){Zheng}, {Wang}, {Liu}, {Li}, {Ye}, and {Ren}}]{2019diou}
Zhaohui {Zheng}, Ping {Wang}, Wei {Liu}, Jinze {Li}, Rongguang {Ye}, and Dongwei {Ren}. 2019.
\newblock \href {https://doi.org/10.48550/arXiv.1911.08287} {{Distance-IoU Loss: Faster and Better Learning for Bounding Box Regression}}.
\newblock \emph{arXiv e-prints}, page arXiv:1911.08287.

\bibitem[{{Zhou} et~al.(2017){Zhou}, {Yao}, {Wen}, {Wang}, {Zhou}, {He}, and {Liang}}]{EAST}
Xinyu {Zhou}, Cong {Yao}, He~{Wen}, Yuzhi {Wang}, Shuchang {Zhou}, Weiran {He}, and Jiajun {Liang}. 2017.
\newblock \href {https://doi.org/10.48550/arXiv.1704.03155} {{EAST: An Efficient and Accurate Scene Text Detector}}.
\newblock \emph{arXiv e-prints}, page arXiv:1704.03155.

\bibitem[{{Zhu} et~al.(2022){Zhu}, {Sokhandan}, {Yang}, {Martin}, and {Sathyanarayana}}]{DocBed}
Wenzhen {Zhu}, Negin {Sokhandan}, Guang {Yang}, Sujitha {Martin}, and Suchitra {Sathyanarayana}. 2022.
\newblock \href {https://doi.org/10.48550/arXiv.2202.01414} {{DocBed: A Multi-Stage OCR Solution for Documents with Complex Layouts}}.
\newblock \emph{arXiv e-prints}, page arXiv:2202.01414.

\end{thebibliography}

\appendix

\onecolumn
\clearpage
\section{Dataset Examples}
\setcounter{table}{0}   
\setcounter{figure}{0}
\renewcommand{\thetable}{A\arabic{table}}
\renewcommand{\thefigure}{A\arabic{figure}}

To the best of our knowledge, this is the first paired dataset containing markup-formatted document contents along with corresponding bounding boxes.
What makes our dataset distinguished from existing ones is that our bounding boxes covers all visible mathematical symbols, such as $\sum$, $\left\langle \right\rangle$ and $\theta^{\alpha}$.

\begin{figure*}[htb]
    \centering 
    \subfigure{
        \label{Fig.sub.1}
        \includegraphics[width=0.5\textwidth]{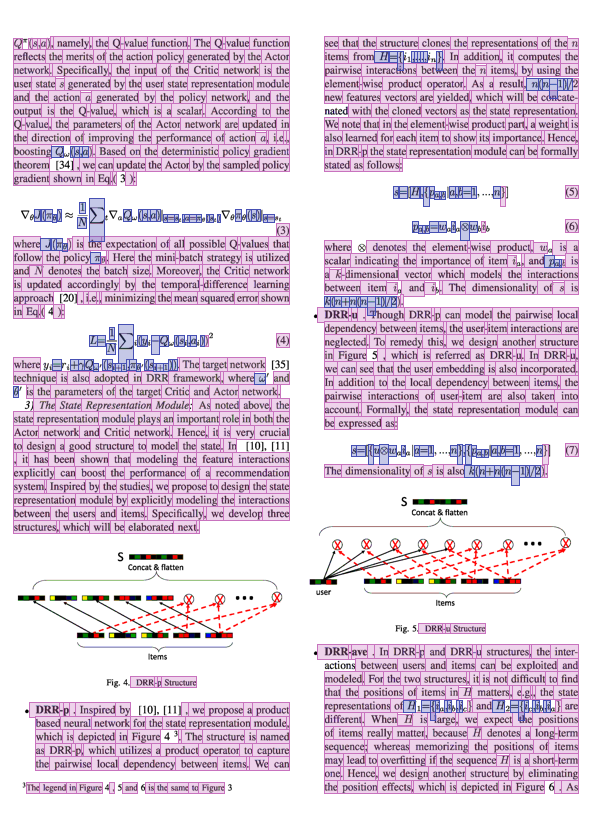}}\subfigure{
        \label{Fig.sub.2}
    \includegraphics[width=0.5\textwidth]{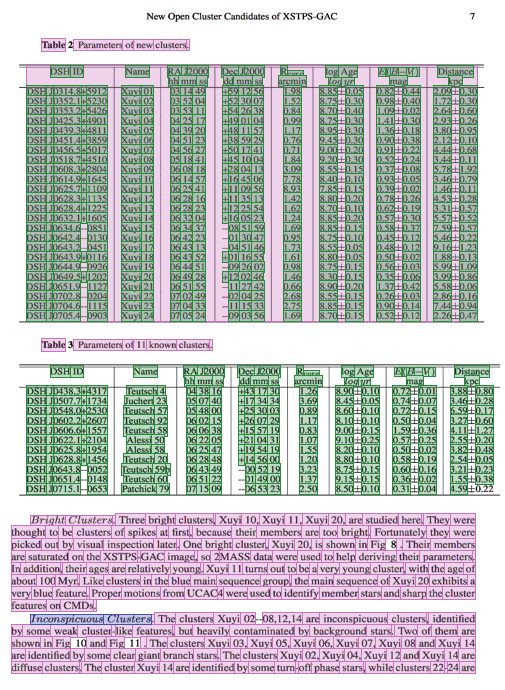}}
    \caption{Dataset example. Bounding boxes of texts are highlighted in pink, mathematical expressions in blue, and tables in green.\\ \\ }
    \label{fig:Dataset}
\end{figure*}

\section{Output Examples}\label{appendix:examples}
\setcounter{table}{0}   
\setcounter{figure}{0}
\renewcommand{\thetable}{B\arabic{table}}
\renewcommand{\thefigure}{B\arabic{figure}}

\begin{figure*}[hbt]
    \centering 
    \subfigure{
        \label{Fig.sub.1}
        \includegraphics[width=0.33\linewidth,height=0.44\linewidth,frame]{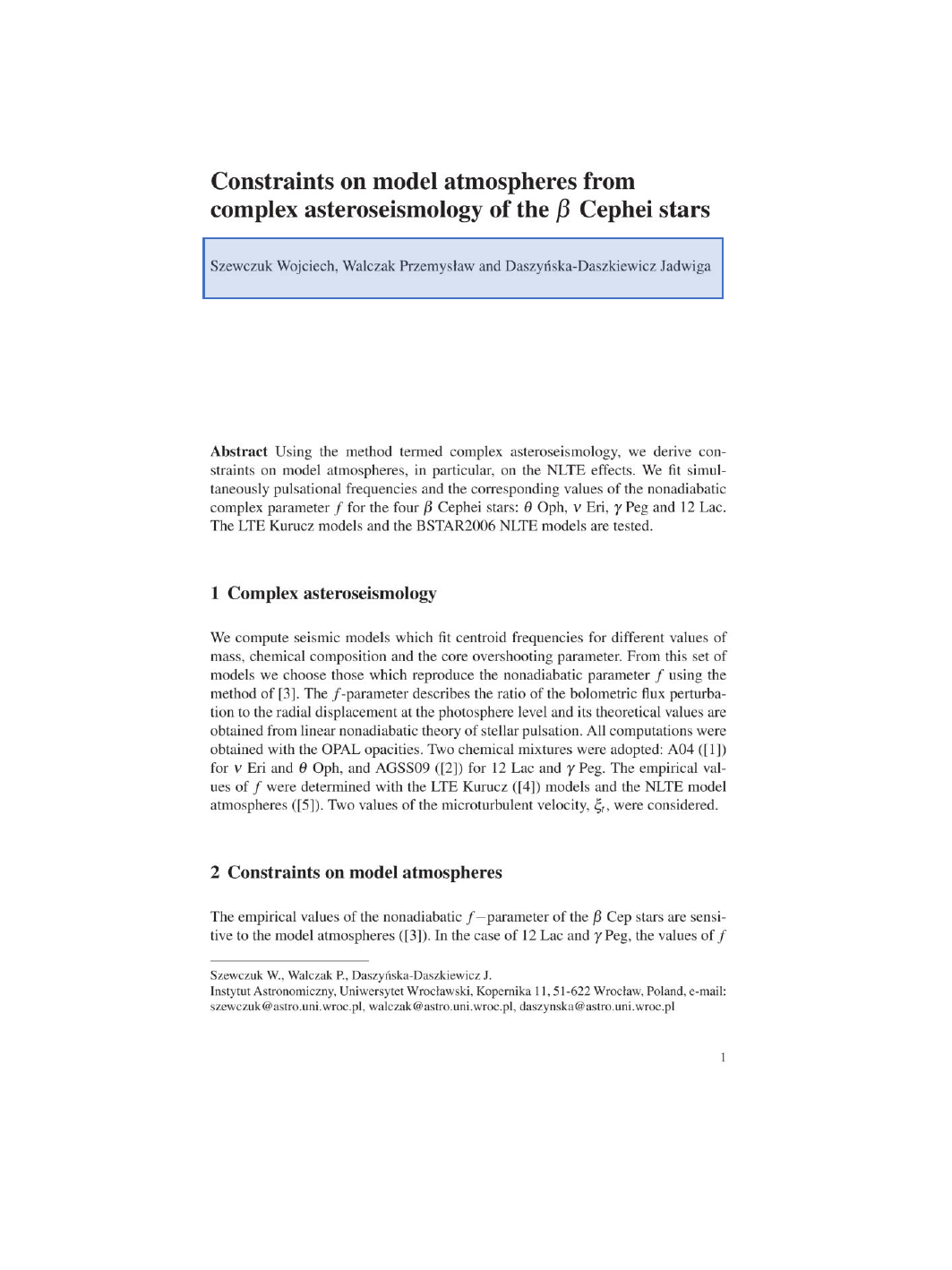}}\subfigure{
        \label{Fig.sub.2}
        \includegraphics[width=0.33\linewidth,height=0.44\linewidth,frame]{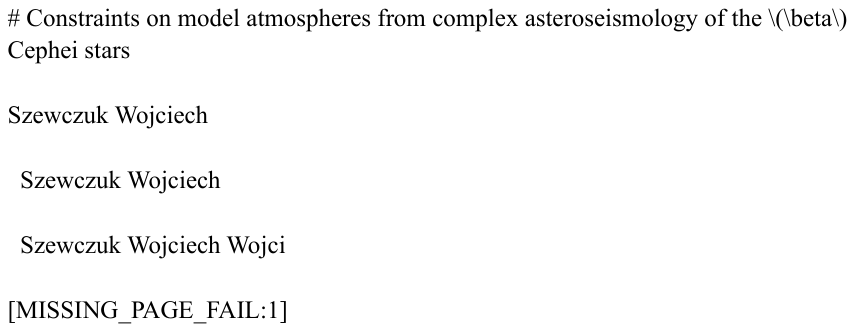}}\subfigure{
        \label{Fig.sub.1}
        \includegraphics[width=0.33\linewidth,height=0.44\linewidth,frame]{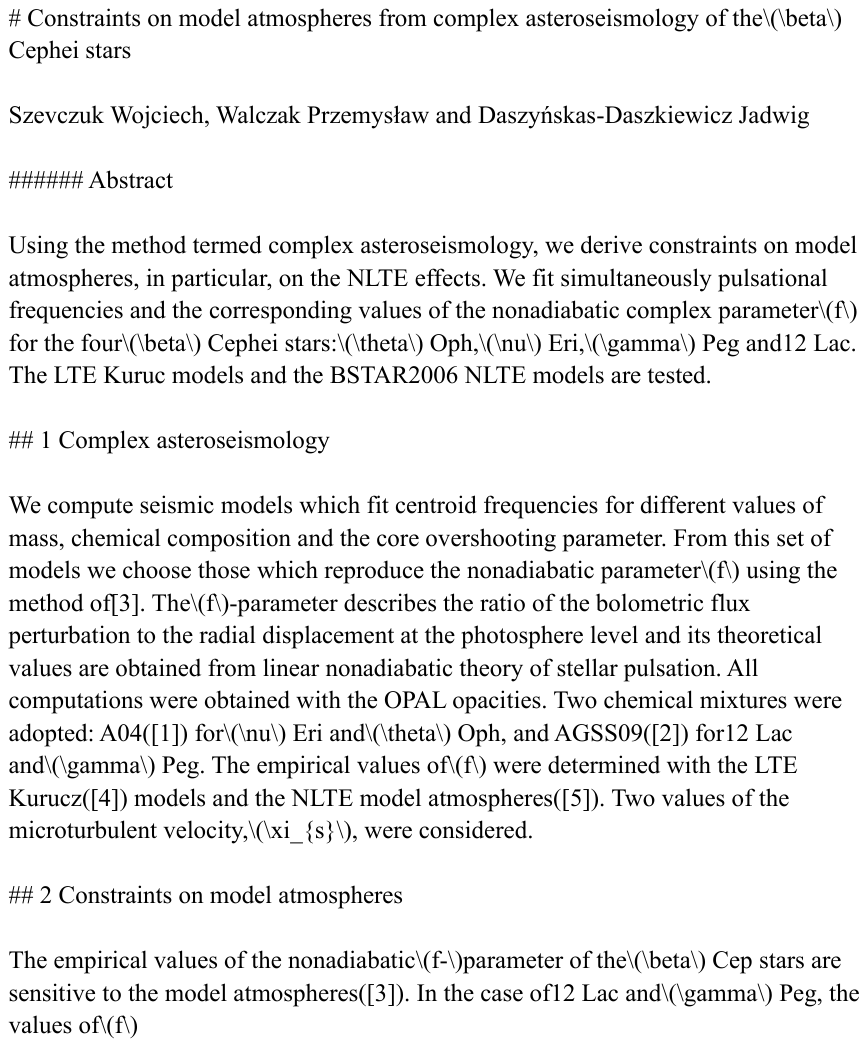}}
    \subfigure{
        \label{Fig.sub.1}
        \includegraphics[width=0.33\linewidth,height=0.44\linewidth,frame]{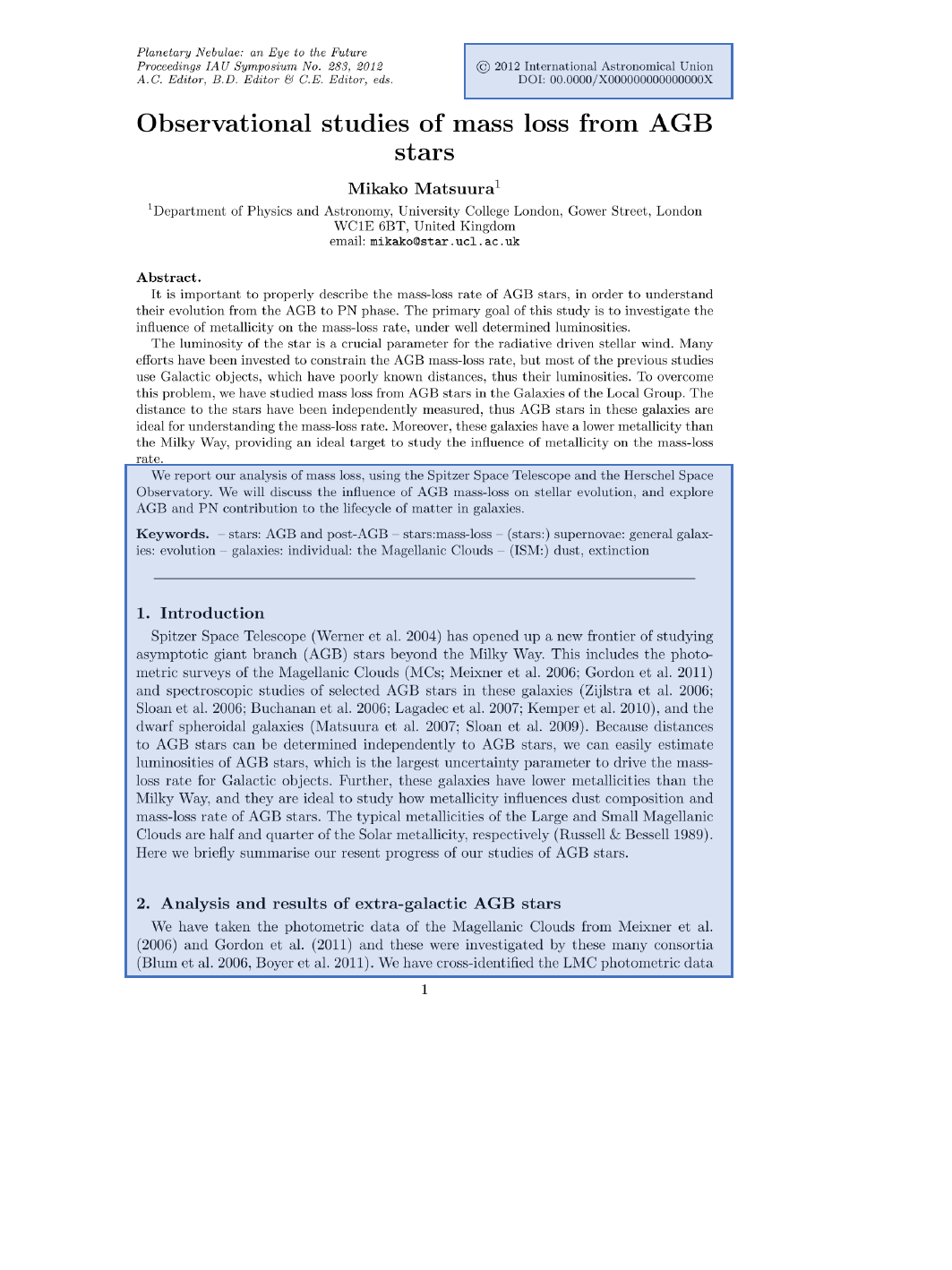}}\subfigure{
        \label{Fig.sub.2}
        \includegraphics[width=0.33\linewidth,height=0.44\linewidth,frame]{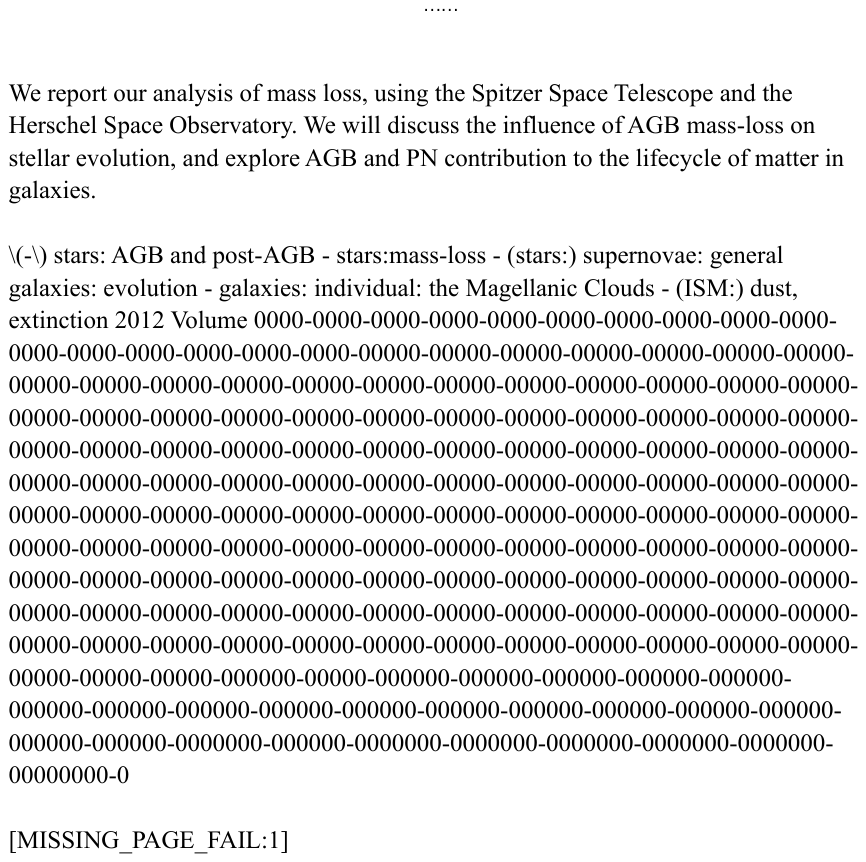}}\subfigure{
        \label{Fig.sub.1}
        \includegraphics[width=0.33\linewidth,height=0.44\linewidth,frame]{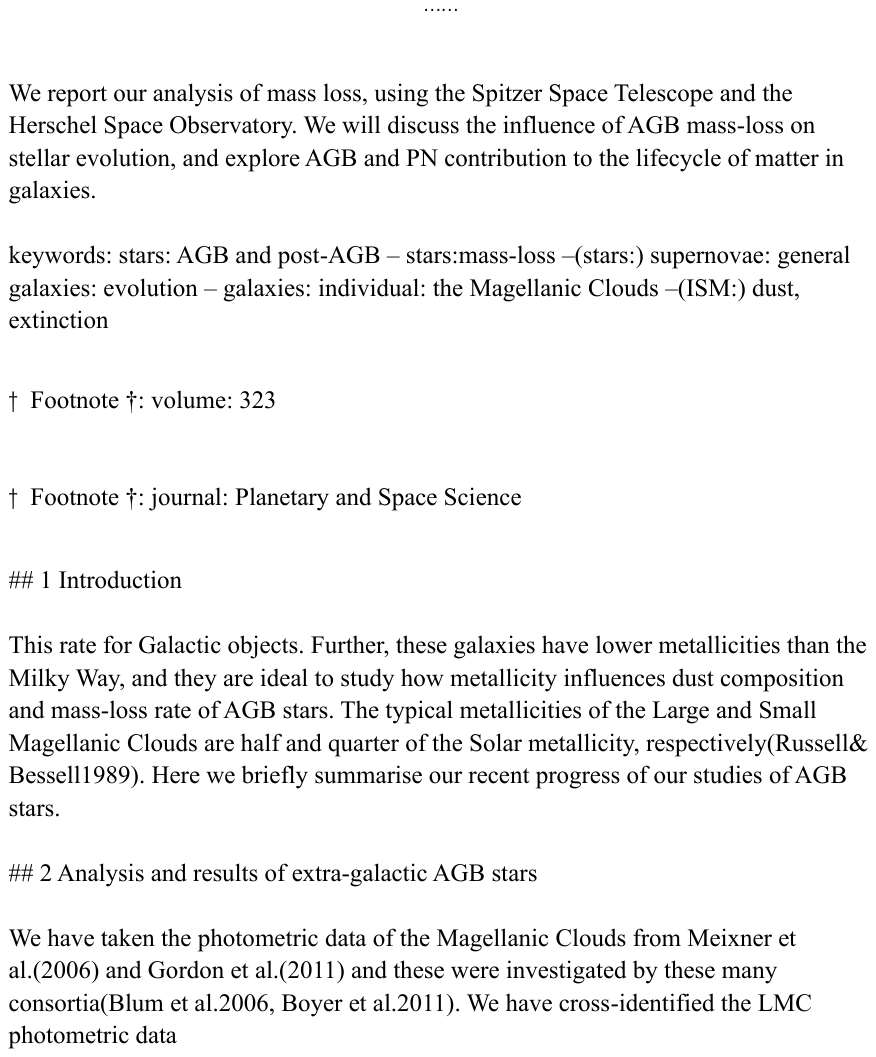}}
    \subfigure{
        \label{Fig.sub.1}
        \includegraphics[width=0.33\linewidth,height=0.44\linewidth,frame]{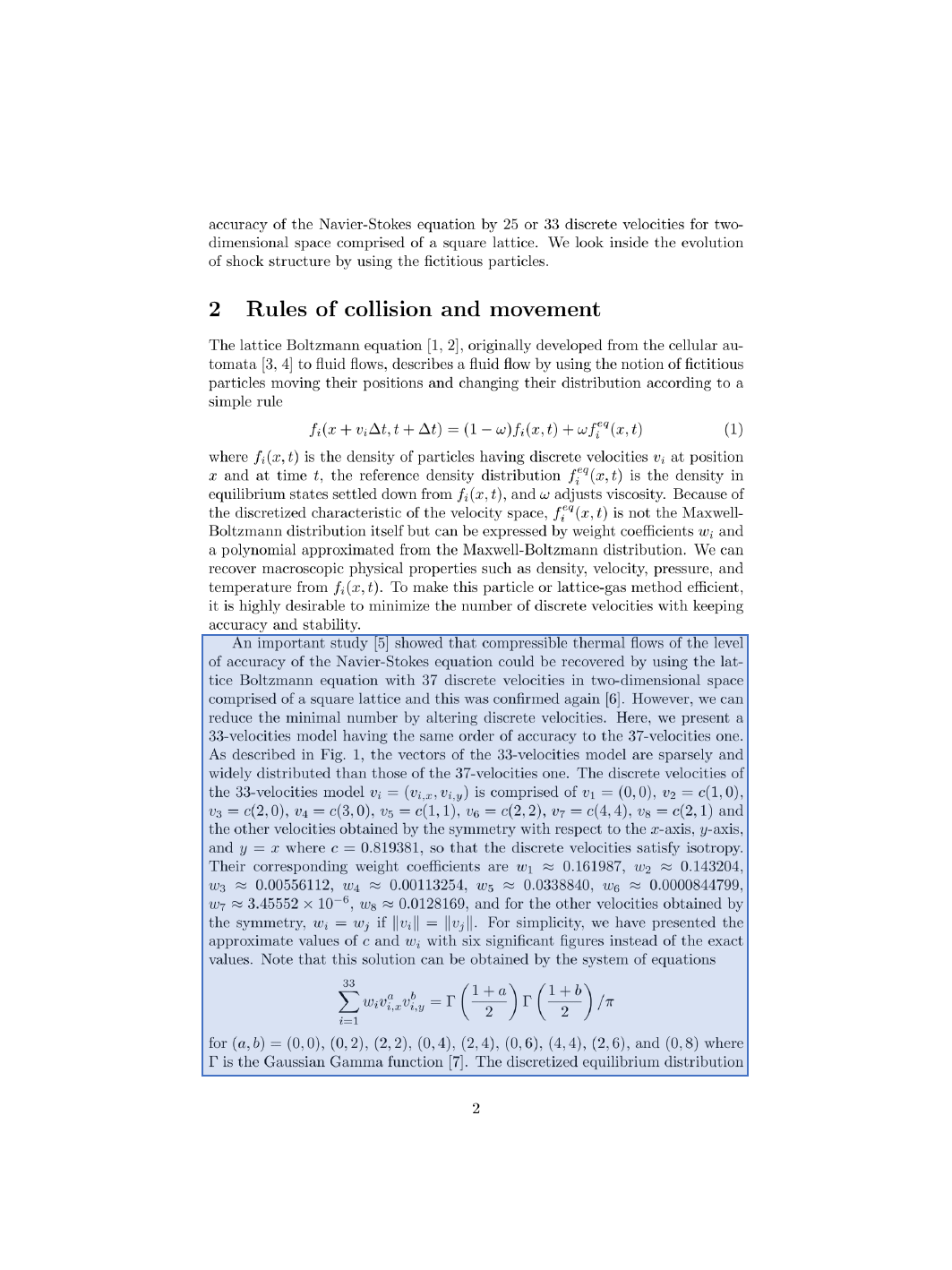}}\subfigure{
        \label{Fig.sub.2}
        \includegraphics[width=0.33\linewidth,height=0.44\linewidth,frame]{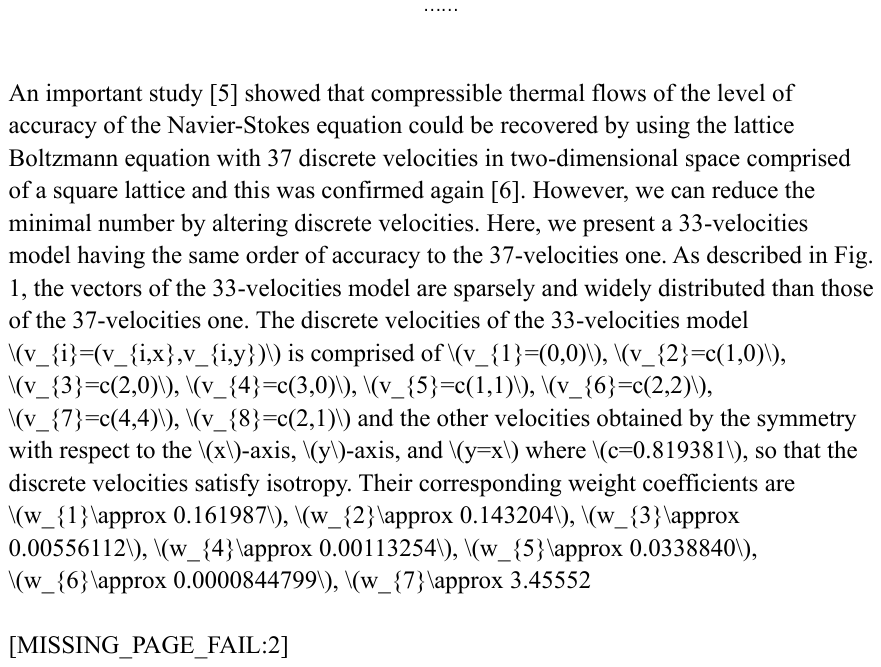}}\subfigure{
        \label{Fig.sub.1}
        \includegraphics[width=0.33\linewidth,height=0.44\linewidth,frame]{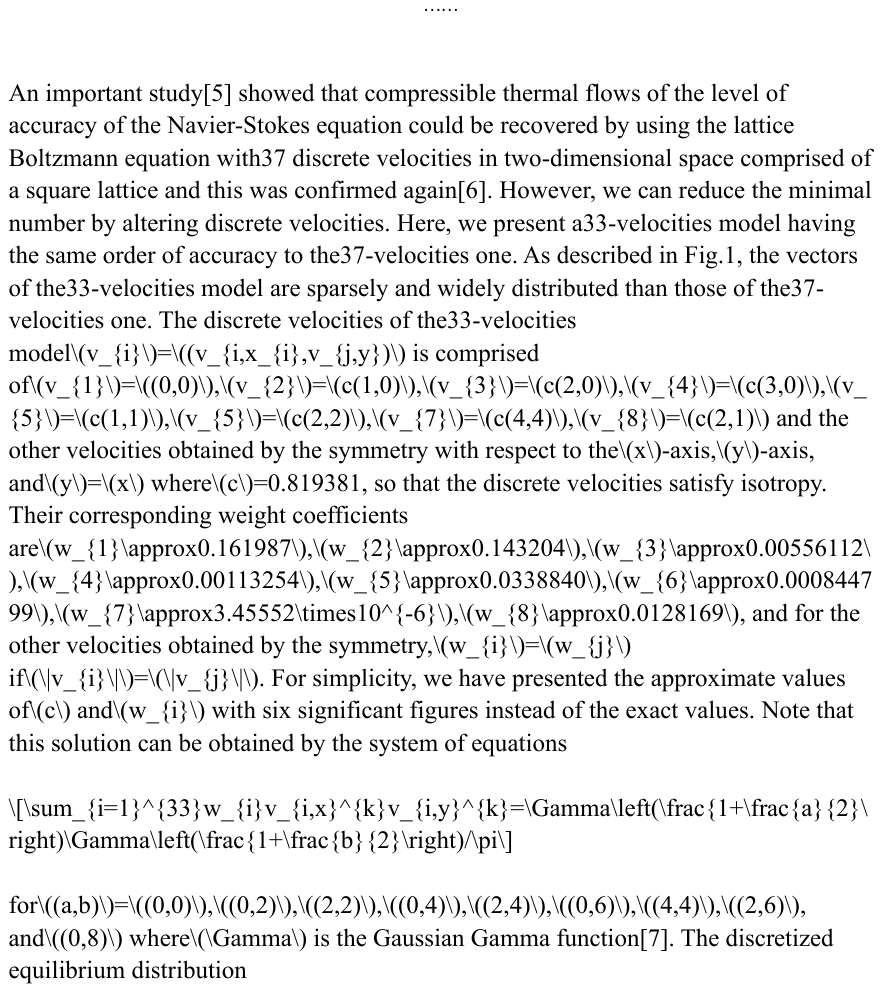}}
    \caption{Examples of pages that Nougat failed to convert but LOCR succeeded. Left: Original PDF pages, with failed parts highlighted in blue. Medium: Markdown output by Nougat. Right: Markdown output by LOCR.}
    \label{fig:mmd result}
\end{figure*}

In Figure~\ref{fig:mmd result}, we compared the output of LOCR and that of Nougat in Markdown format, together with the original PDF pages. Compared with Nougat, LOCR successfully handled the repetition problem. The corresponding part in PDF is highlighted in blue. 

\begin{figure*}[b]
    \centering 
    \subfigure{
        \label{Fig.sub.1}
        \includegraphics[width=0.5\linewidth,height=0.68\linewidth,frame]{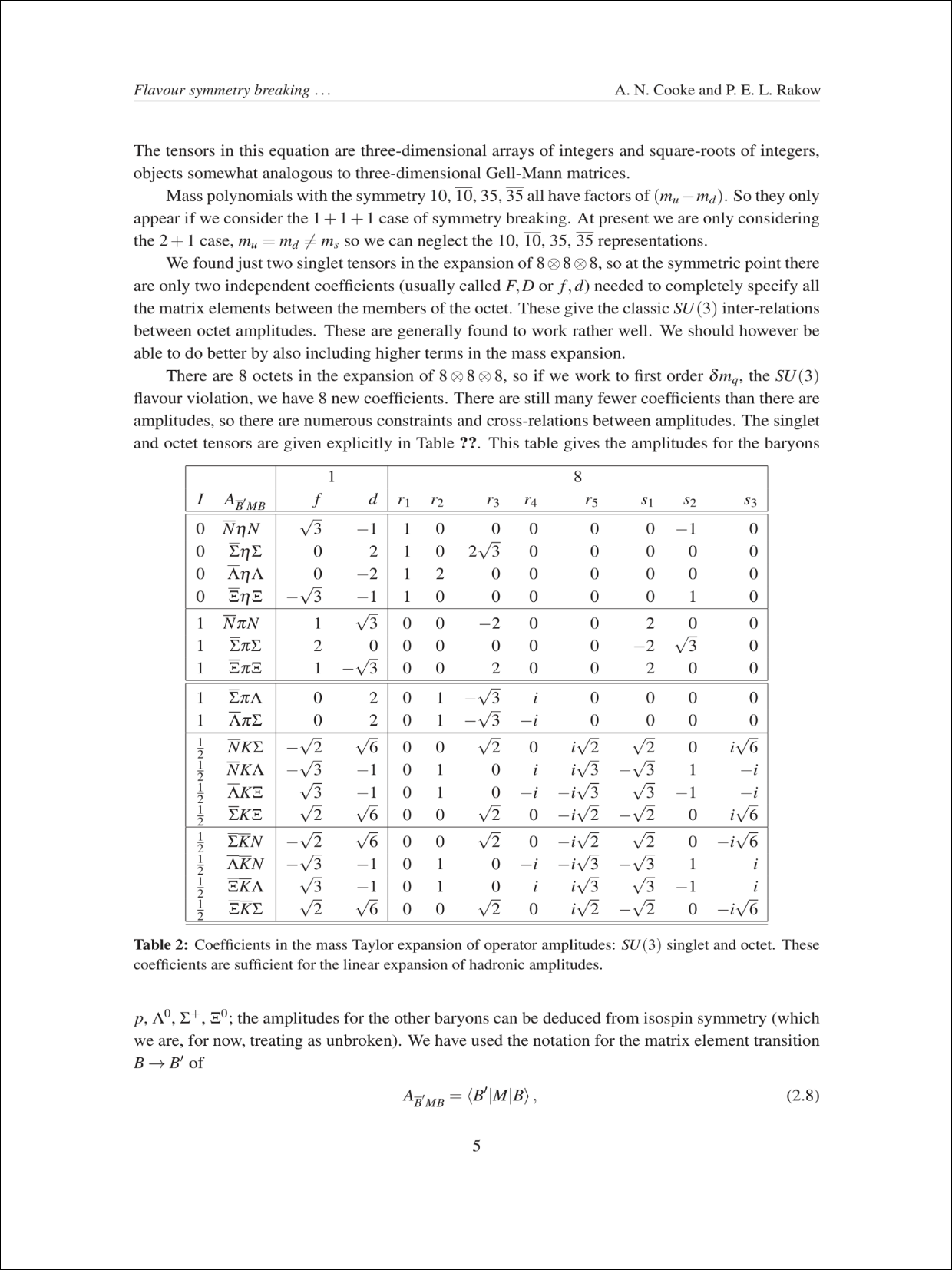}}\subfigure{
        \label{Fig.sub.2}
        \includegraphics[width=0.5\linewidth,,height=0.68\linewidth,frame]{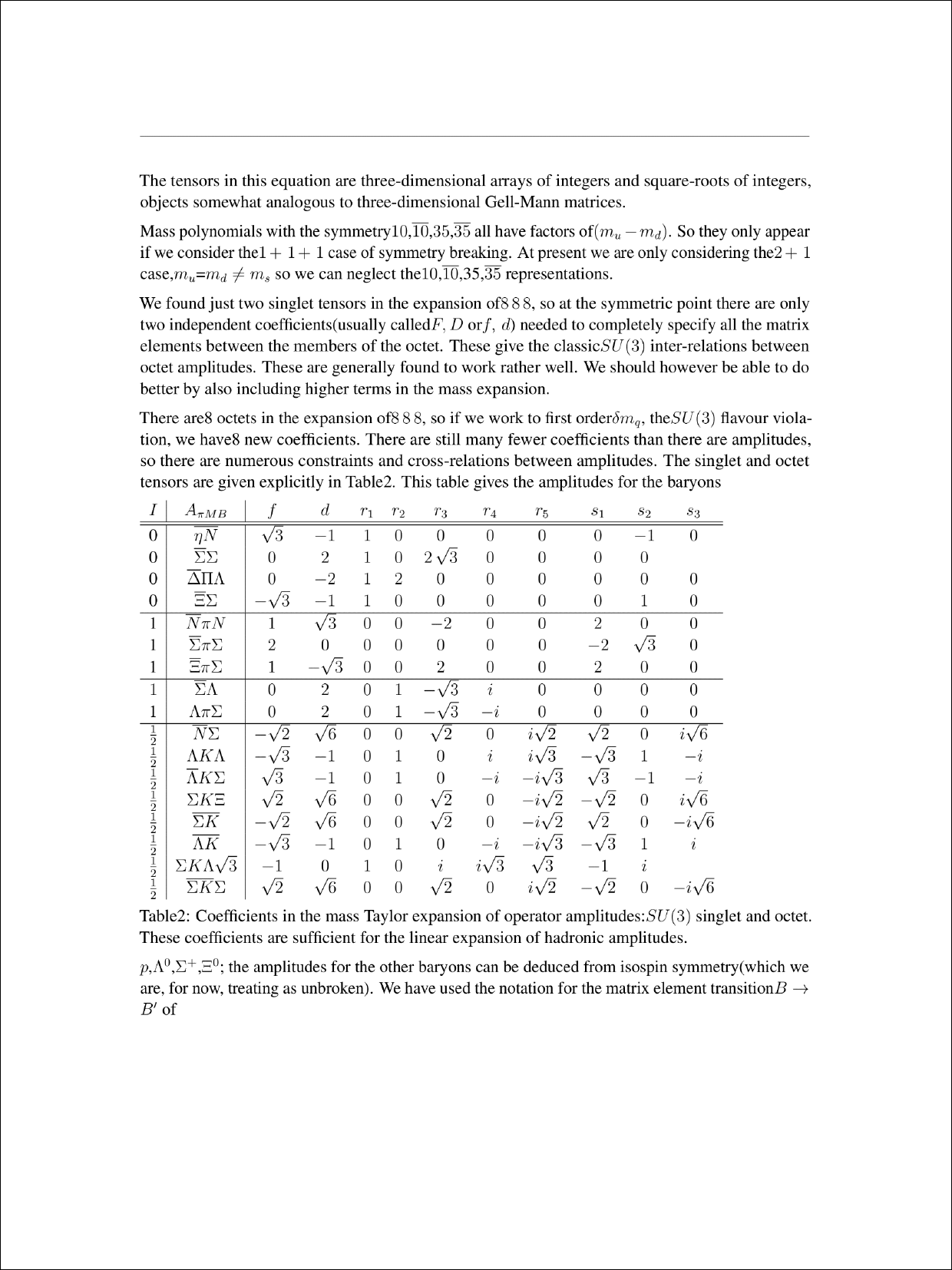}}
        \subfigure{
        \label{Fig.sub.1}
        \includegraphics[width=0.5\linewidth,height=0.68\linewidth,frame]{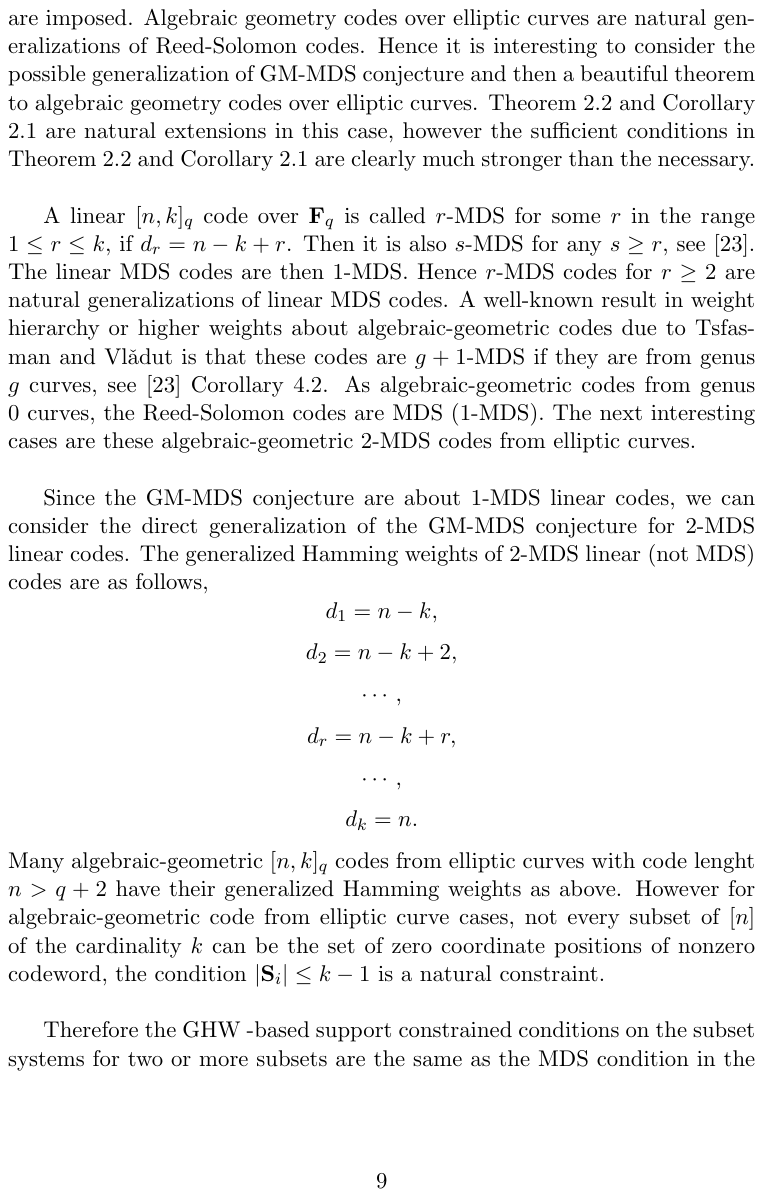}}\subfigure{
        \label{Fig.sub.2}
        \includegraphics[width=0.5\linewidth,height=0.68\linewidth,frame]{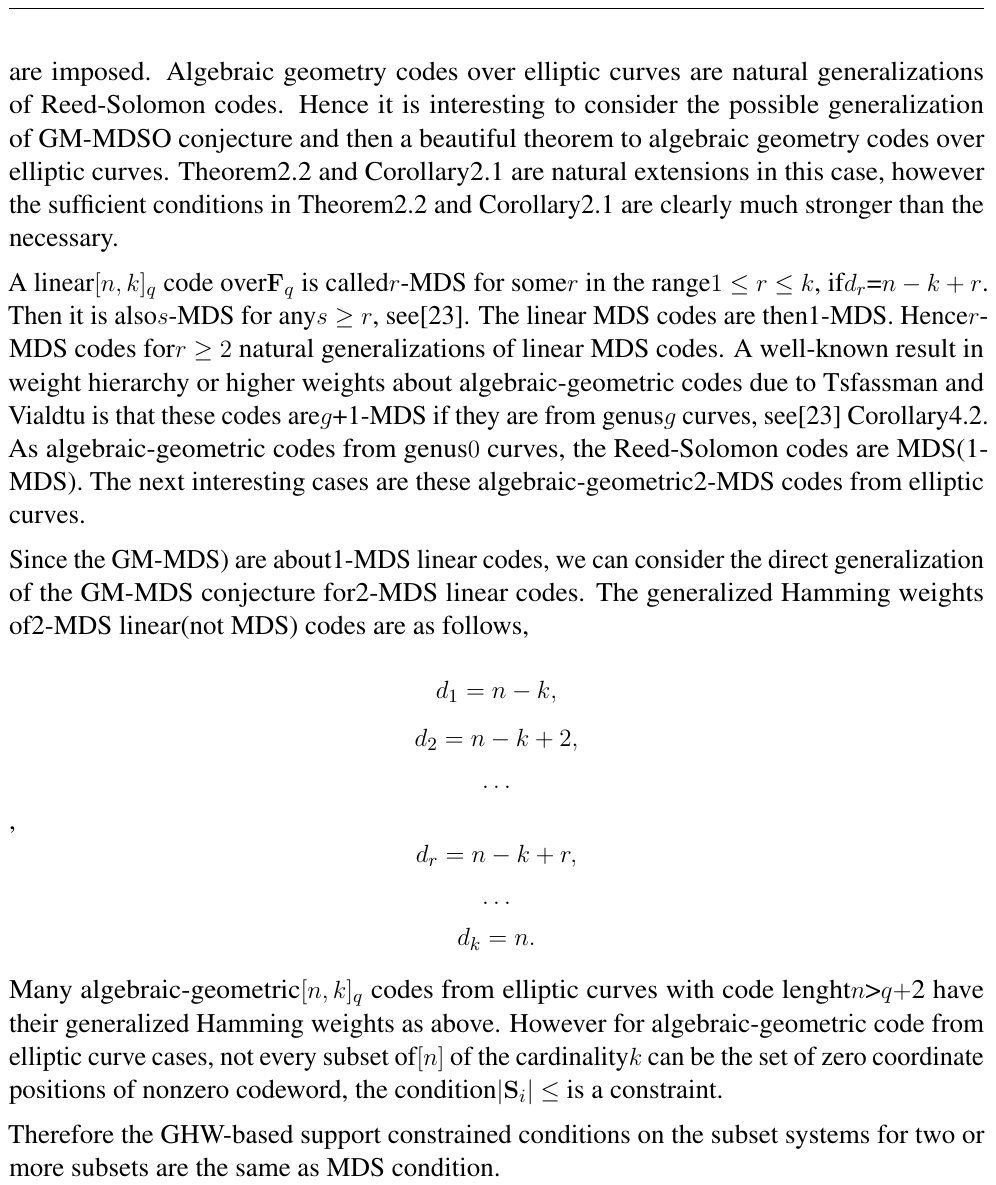}}
    \caption{Examples of our model output. Left: Origin image of document page with tables and equations. Right: Model output converted to Markdown and rendered back into a PDF.}
    \label{fig:recompiled examples}
\end{figure*}

As a more clear illustration, Figure~\ref{fig:recompiled examples} shows the output of LOCR recompiled into PDF format.

Figure~\ref{fig:box prediction} shows the visualization of bounding boxes predicted by position detection head. LOCR predicts bounding boxes with high accuracy not only for plain texts, but also for figure captions, mathematical symbols and tables.

\renewcommand{\dblfloatpagefraction}{.9}

\begin{figure*}
    \centering 
    \subfigure[Origin page with figures]{
        \label{Fig.sub.1}
        \includegraphics[width=0.5\textwidth]{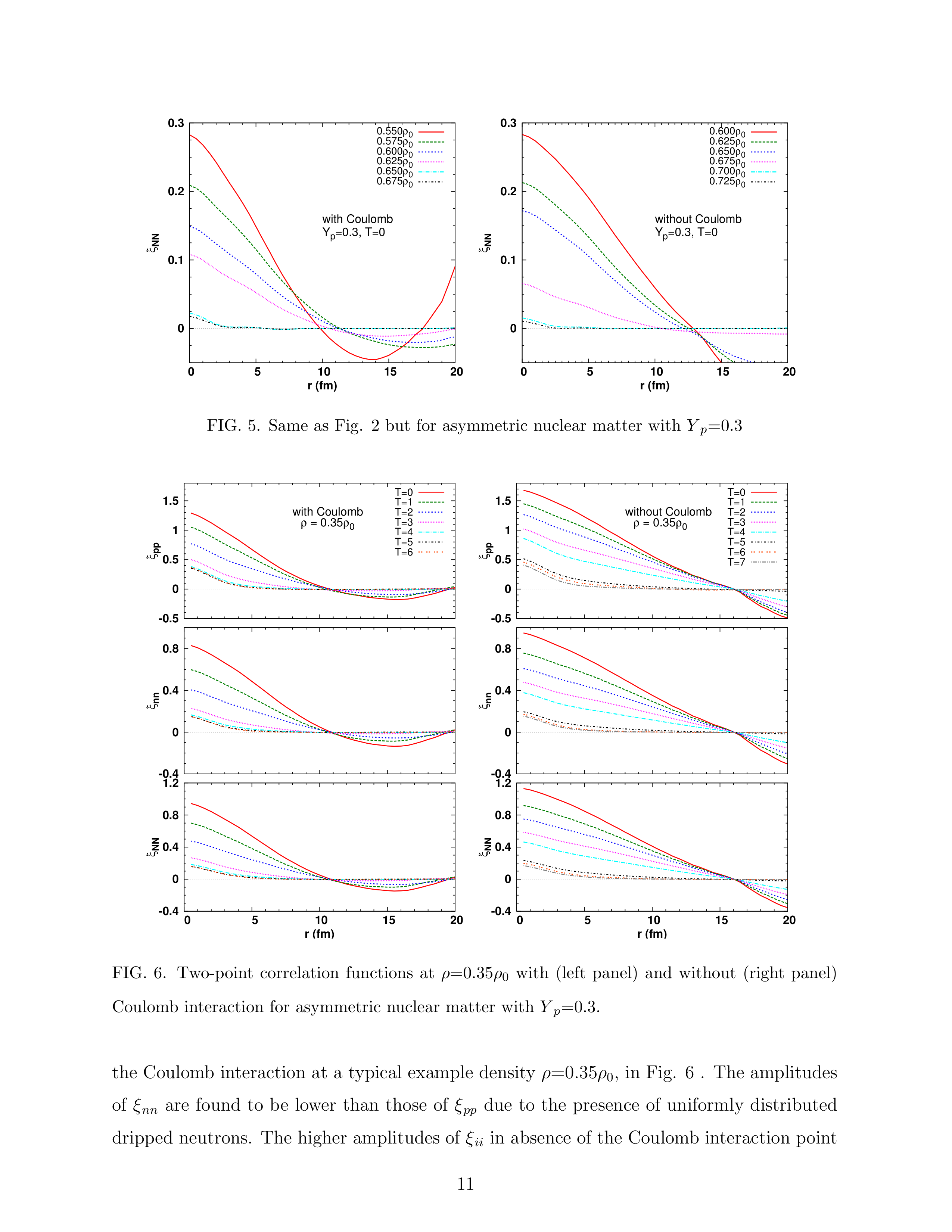}}\subfigure[Result]{
        \label{Fig.sub.2}
    \includegraphics[width=0.5\textwidth]{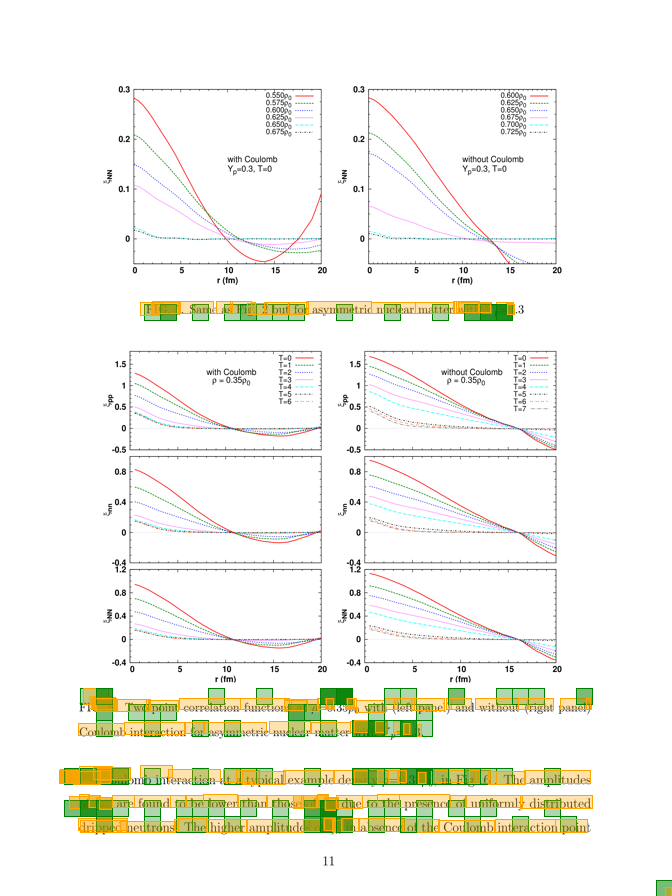}}
    \subfigure[Origin page with mathematical formulas]{
        \label{Fig.sub.1}
        \includegraphics[width=0.5\textwidth]{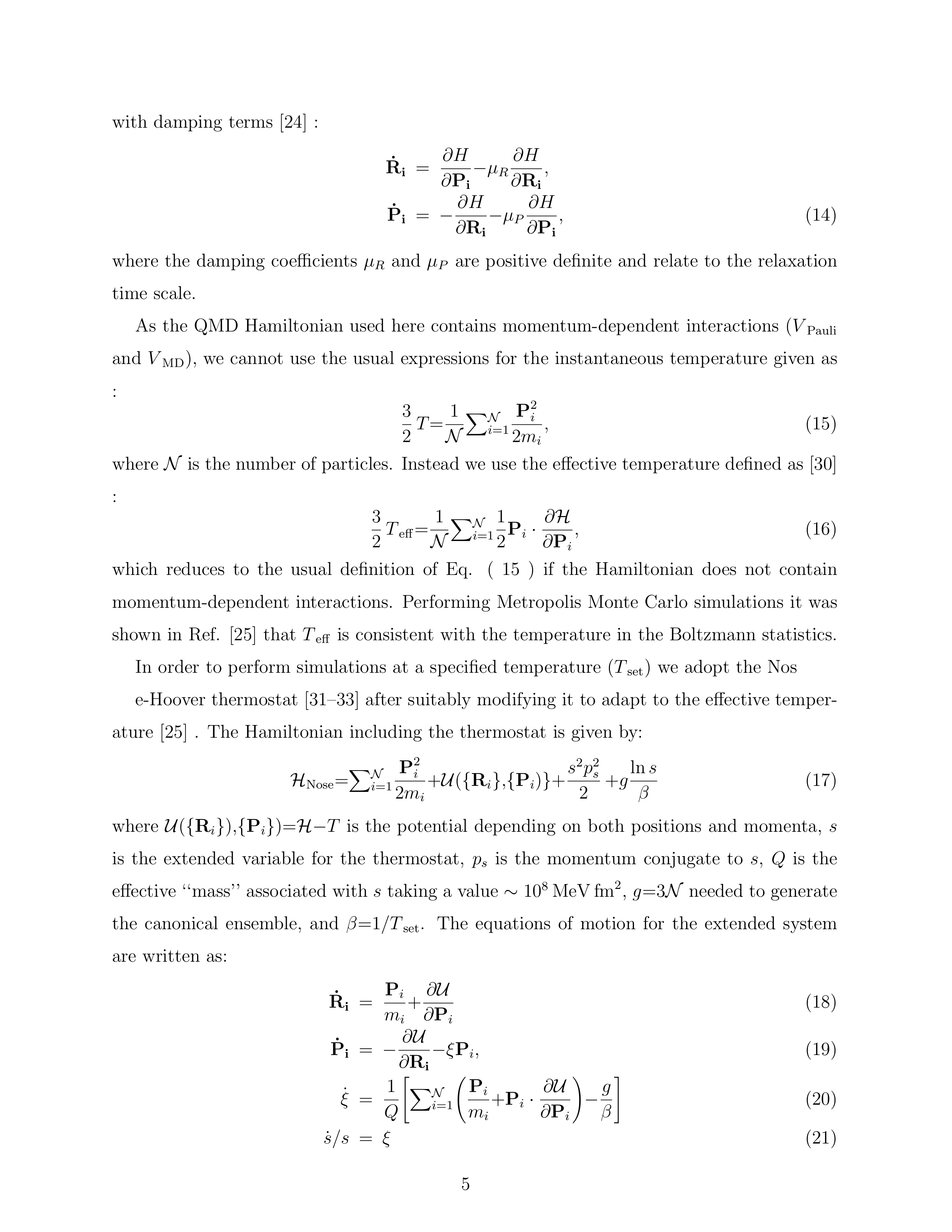}}\subfigure[Result]{
        \label{Fig.sub.2}
    \includegraphics[width=0.5\textwidth]{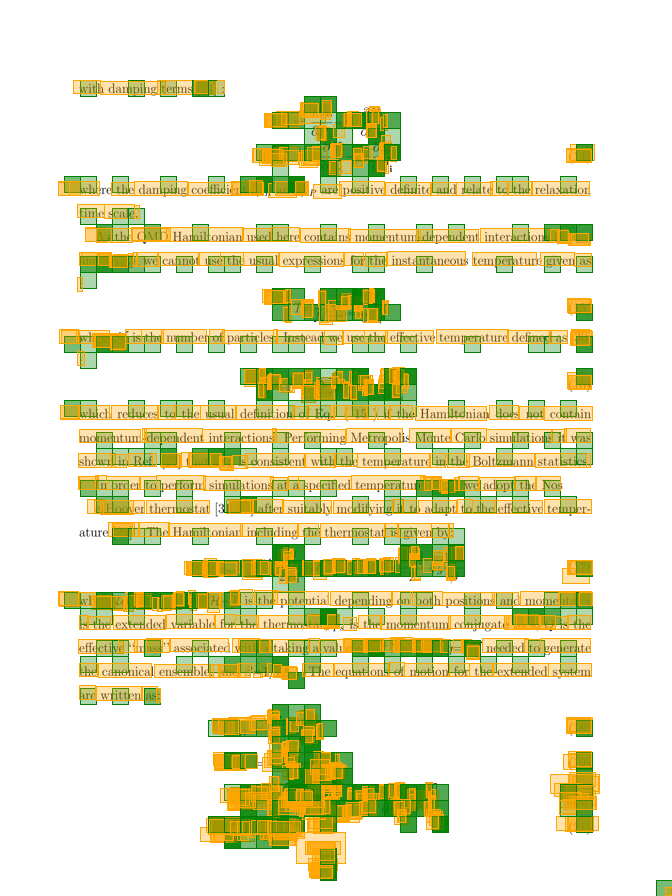}}
    \label{fig:box prediction}
\end{figure*}

\begin{figure*}
    \centering 
    \ContinuedFloat
    \addtocounter{figure}{1}
    \subfigure[Origin page with tables]{
        \label{Fig.sub.1}
        \includegraphics[width=0.5\textwidth]{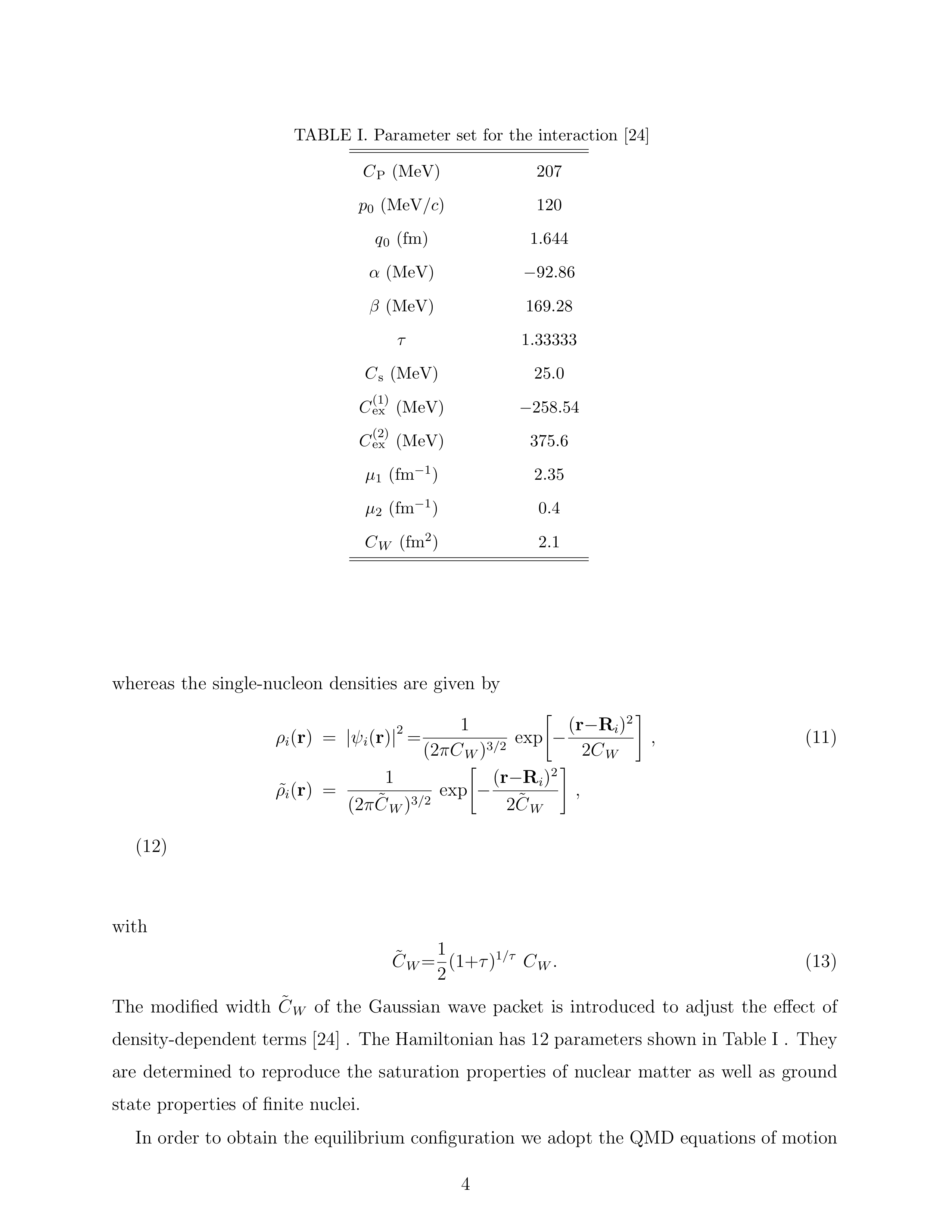}}\subfigure[Result]{
        \label{Fig.sub.2}
    \includegraphics[width=0.5\textwidth]{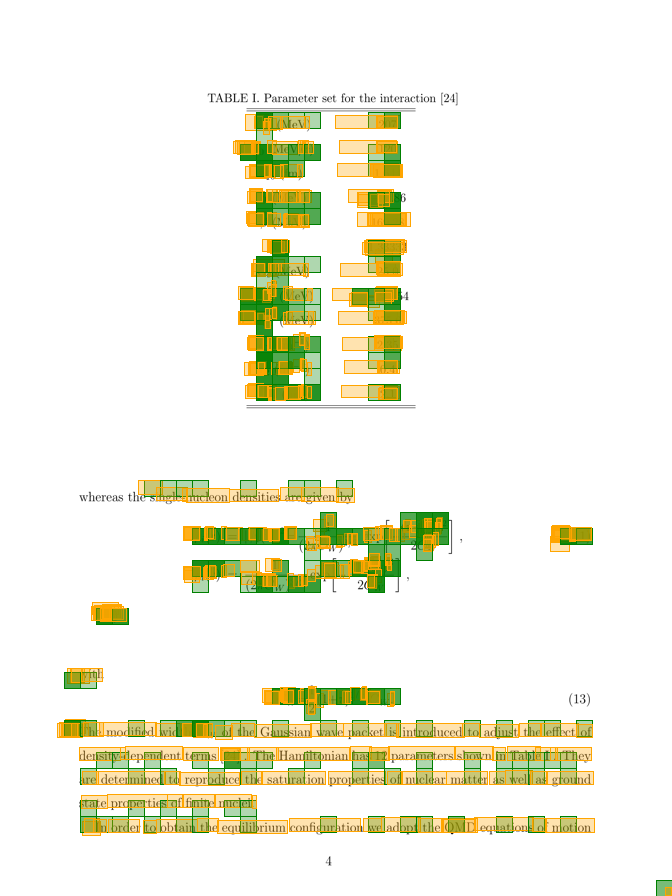}}
    \subfigure[Origin page with references]{
        \label{Fig.sub.1}
        \includegraphics[width=0.5\textwidth]{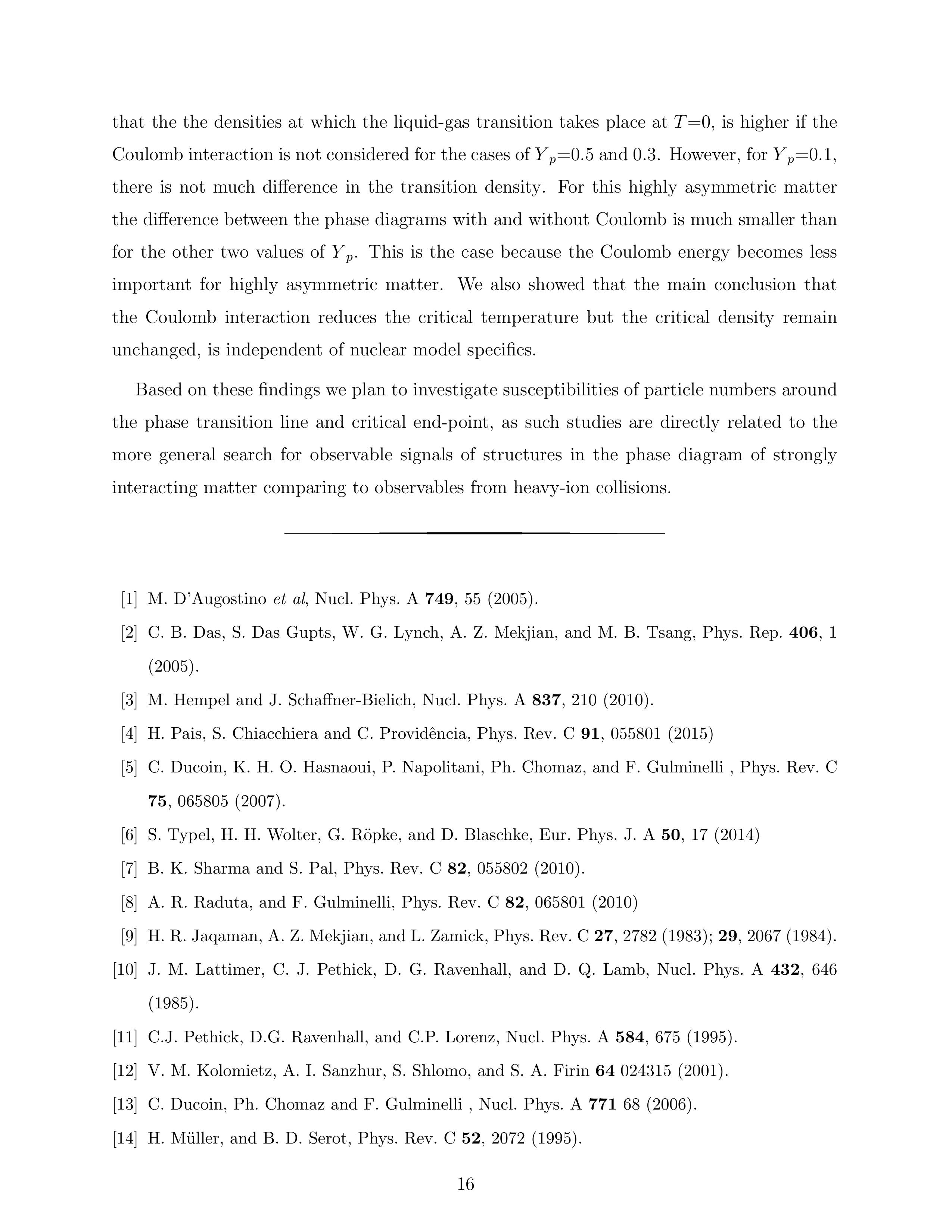}}\subfigure[Result]{
        \label{Fig.sub.2}
    \includegraphics[width=0.5\textwidth]{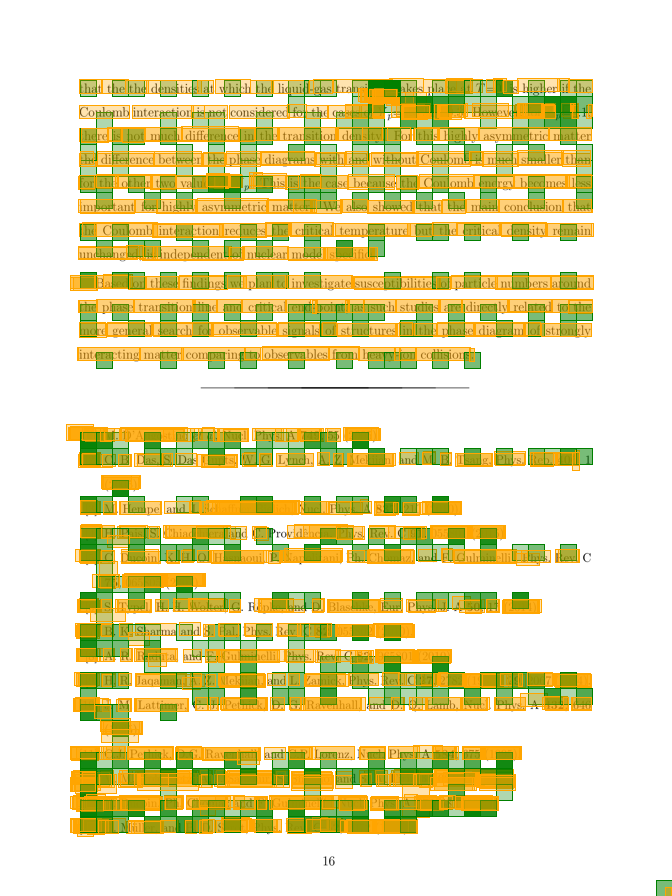}}
    \caption{Example of position prediction. Green box: Rough result of grid classification. Yellow: Final result of box regression.}
    \label{fig:box prediction}
\end{figure*}

\clearpage
\section{Interactive Mode}\label{appendix:interaction mode}
\setcounter{table}{0} 
\setcounter{figure}{0}
\renewcommand{\thetable}{C\arabic{table}}
\renewcommand{\thefigure}{C\arabic{figure}}
\renewcommand{\dblfloatpagefraction}{.9}

\begin{figure*}[hbtp]
    \centering 
    \subfigure[A case model predicting wrong position]{
        \label{subfig.1}
        \includegraphics[width=0.5\textwidth]{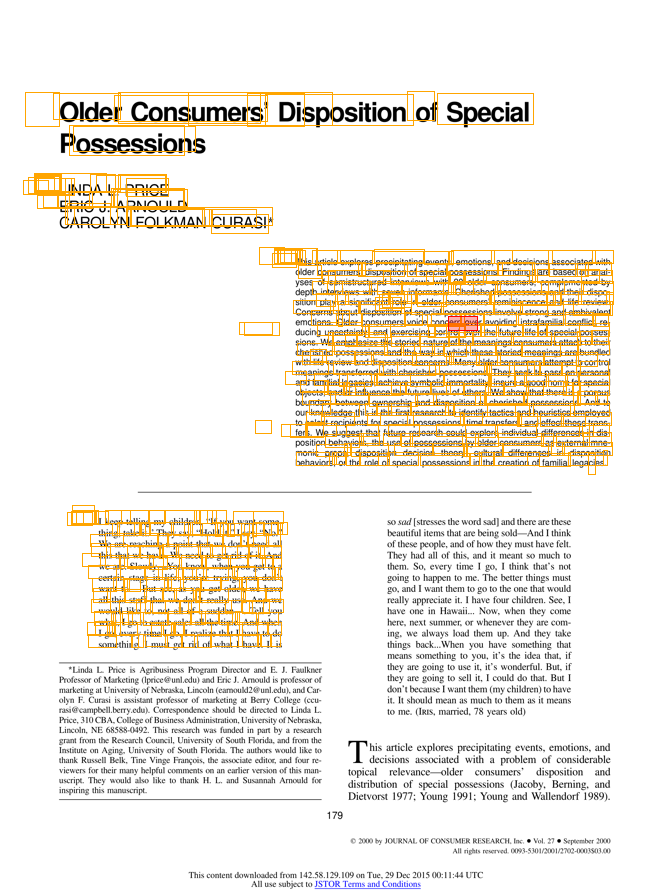}}\subfigure[Result]{
        \label{subfig.2}
    \includegraphics[width=0.5\textwidth]{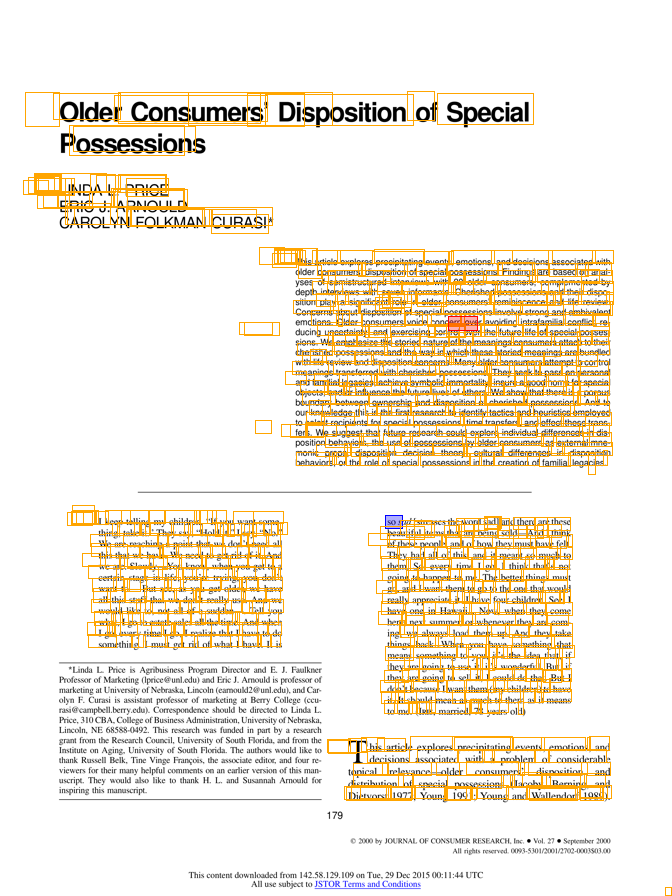}}
    \caption{Visualization of interaction on out-of-domain documents. Red box: Wrong position. Blue box: Human prompt input.}
    \label{fig:interactive mode}
\end{figure*}

Figure~\ref{fig:interactive mode} shows the interactive process with human intervention. The orange bounding boxes denote the areas that have been scanned by the model. The model predicted a low confidence score when it decoded to the position shown in \ref{subfig.1}, with the incorrectly predicted position highlighted in red. In \ref{subfig.2}, human gave a box prompt highlighted in blue and the model output the subsequent contents smoothly and correctly.

\clearpage
\section{Statistics of Test Documents} \label{appendix:statistics of test documents}
\setcounter{table}{0} 
\setcounter{figure}{0}
\renewcommand{\thetable}{D\arabic{table}}
\renewcommand{\thefigure}{D\arabic{figure}}

\begin{figure*}[htbp]
    \centering
    \includegraphics[width=1\linewidth]{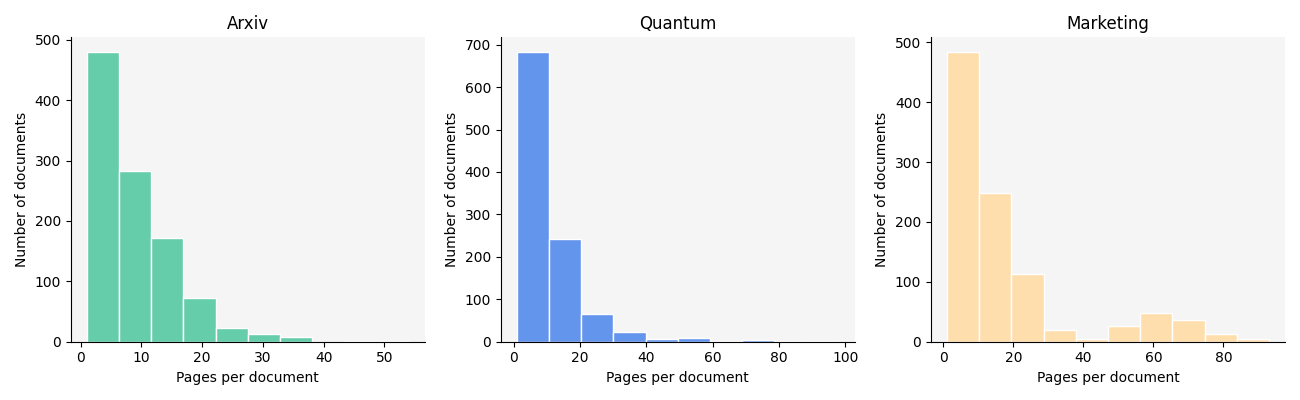}
    \caption{Histograms of the number of pages per document in each repetition test set. }
    \label{fig:histogram}
\end{figure*}

As a complementary illustration for Table~\ref{repetition metrics}, we show the histograms of the number of pages per document in Figure~\ref{fig:histogram}. Consistent with the conclusion in Table~\ref{repetition metrics}, when counting in document number, domains with more pages per document, such as marketing, have a higher generation failure rate.

\section{A case when Nougat gets trapped into repetition} \label{appendix:nougat repetition}
\setcounter{table}{0} 
\setcounter{figure}{0}
\renewcommand{\thetable}{E\arabic{table}}
\renewcommand{\thefigure}{E\arabic{figure}}

\begin{figure*}[hbtp]
    \centering 
    \subfigure[Correct attentions for the authors.]{
        \label{subfig.1}
        \includegraphics[width=0.33\textwidth]{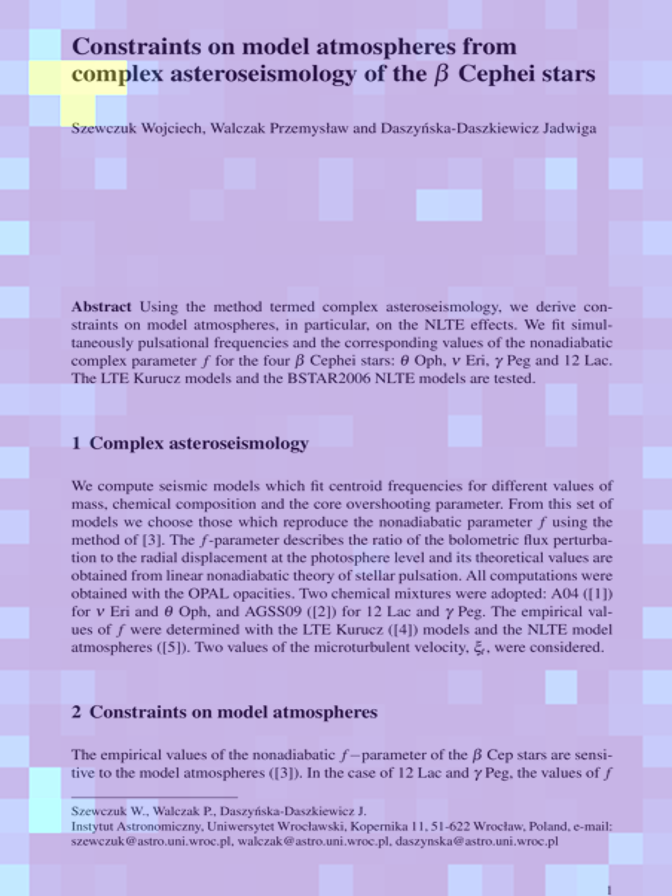}}\subfigure[Correct attentions for the footnote]{
        \label{subfig.2}
    \includegraphics[width=0.33\textwidth]{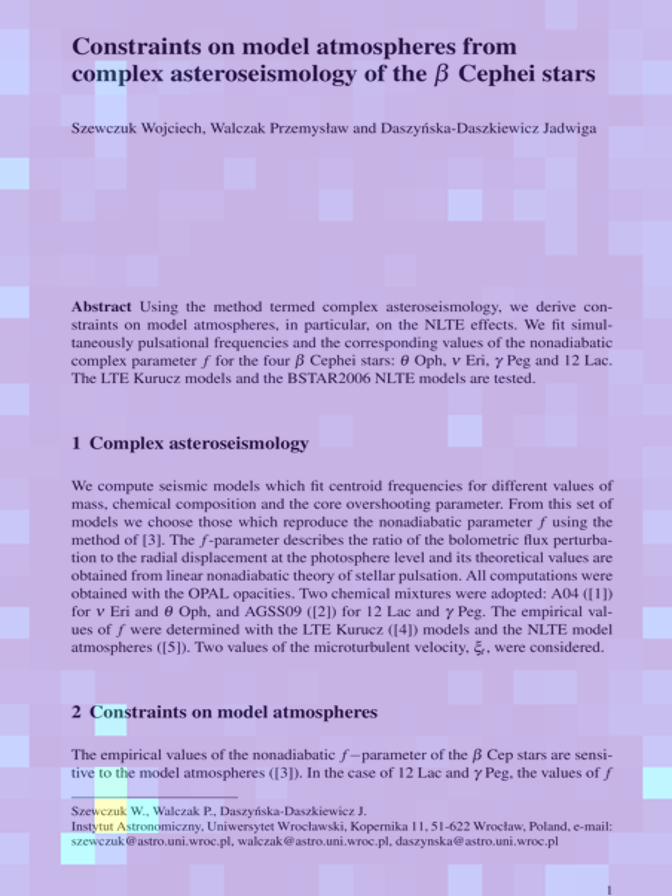}}\subfigure[Incorrect attentions when repetition.]{
        \label{subfig.2}
    \includegraphics[width=0.33\textwidth]{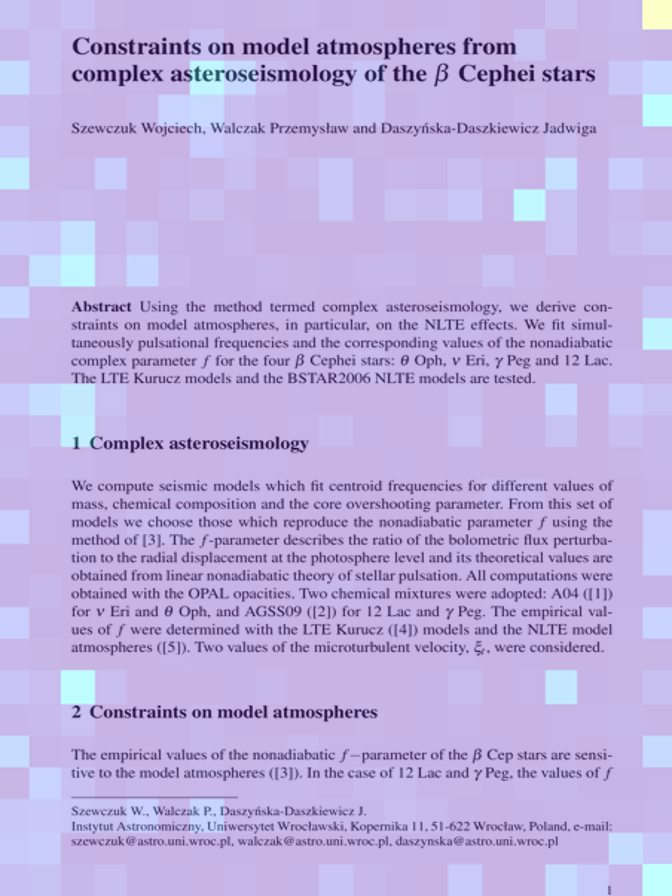}}
    \caption{The heatmap of cross-attention of Nougat, in which yellow denotes larger attention scores and purple denotes smaller scores. Left: Cross-attention scores when Nougat decoded to the name of the first author. Medium: Cross-attention scores when Nougat tried to decode the footnote. Right: Cross-attention scores when Nougat began repetition and failed to find the correct position.}
    \label{fig:nougat repetition}
\end{figure*}

Figure~\ref{fig:nougat repetition} shows a case when nougat got trapped into repetition. After decoding the name of the first author, Nougat tried to find the correlation between the footnote and the authors but failed. The heatmap of cross-attenions ended with cycling through the three subfigures and the output ended with repeating the name "Szewczuk Wojciech Szewczuk Wojciech Szewczuk Wojciech Wojci". The original PDF page, the output of Nougat and that of LOCR is shown in Figure~\ref{fig:mmd result}.

\end{document}